\documentclass{ecai} 

\usepackage{latexsym}
\usepackage{amssymb}
\usepackage{amsmath}
\usepackage{amsthm}
\usepackage{booktabs}
\usepackage{enumitem}
\usepackage{graphicx}
\usepackage{color}
\usepackage{pifont}
\usepackage{multirow}
\usepackage{array}
\usepackage{comment}

\newcommand{\BibTeX}{B\kern-.05em{\sc i\kern-.025em b}\kern-.08em\TeX}

\newcommand{\bx}{\textbf{x}}
\newcommand{\bz}{\textbf{z}}

\newcommand{\be}{\textbf{e}}
\newcommand{\bE}{\textbf{E}}

\newcommand{\bW}{\textbf{W}}

\newcommand{\Dsyn}{\mathcal{D}_{\text{syn}}}

\newcommand{\cmark}{\ding{51}}
\newcommand{\xmark}{\ding{55}}

\begin{document}


\begin{frontmatter}


\paperid{8105}


\title{Exploring Transformer Placement in Variational Autoencoders for Tabular Data Generation}


\author[A]{\fnms{Aníbal}~\snm{Silva}
}
\author[B]{\fnms{Moisés}~\snm{Santos}
}
\author[B]{\fnms{André}~\snm{Restivo}
} 
\author[B]{\fnms{Carlos}~\snm{Soares}
} 

\address[A]{Faculty of Sciences, University of Porto}
\address[B]{Faculty of Engineering, University of Porto}


\begin{abstract}


Tabular data remains a challenging domain for generative models. In particular, the standard Variational Autoencoder (VAE) architecture, typically composed of multilayer perceptrons, struggles to model relationships between features, especially when handling mixed data types. In contrast, Transformers, through their attention mechanism, are better suited for capturing complex feature interactions. In this paper, we empirically investigate the impact of integrating Transformers into different components of a VAE. We conduct experiments on 57 datasets from the OpenML CC18 suite and draw two main conclusions. First, results indicate that positioning Transformers to leverage latent and decoder representations leads to a trade-off between fidelity and diversity. Second, we observe a high similarity between consecutive blocks of a Transformer in all components. In particular, in the decoder, the relationship between the input and output of a Transformer is approximately linear.

\end{abstract}

\end{frontmatter}


\section{Introduction}\label{sec:intro}

Deep Learning has been thoroughly investigated over the last decades and has been successfully applied to various learning tasks. Generative modeling is no exception. Generally, this class of models aims to estimate the underlying distribution of the data. Existing generative flavors encompass variational inference~\cite{Kingma2013AutoEncodingVB, 10.5555/3495724.3496298, pmlr-v37-rezende15}, generative adversarial networks~\cite{NIPS2014_5ca3e9b1} and score-based matching~\cite{10.5555/3454287.3455354}. Despite mostly focusing on data modalities such as image~\cite{Bauer2024ComprehensiveEO} and text~\cite{10.1145/3649449}, there has been a recent surge of interest in generative models for tabular data. The interest lies in generating synthetic data to overcome challenges such as data scarcity, missing-value imputation, and individual privacy-preserving (see e.g.,~\cite{r2024navigatingtabulardatasynthesis} for a thorough review).

\paragraph{Challenges in tabular data generation}

Modeling the joint distribution of tabular data has unique challenges. The main research interest in the generative model is on images, and, usually, the theory behind it assumes a continuous distribution of the data. This is not true for tabular data, which generally presents a mixture of both continuous and discrete variables. Moreover, continuous variables might exhibit several modes, and discrete variables may have a considerable number of categories, imposing additional challenges on the capability of a neural network to learn relationships between these two different types of data adequately. 

\paragraph{Tokenization}

A data point of a tabular dataset is usually represented as a heterogeneous vector composed of numerical and discrete features. One possible solution to overcome this heterogeneity is to embed each feature into an embedding matrix via tokenization~\cite{Gorishniy2021RevisitingDL, Zhang2023MixedTypeTD}. In essence, this transformation linearly projects each feature into a continuous vector.


\paragraph{Transformers}

The Transformer architecture~\cite{10.5555/3295222.3295349} was initially proposed for machine translation and later applied to text generation~\cite{Radford2018ImprovingLU}. Given its unprecedented success, adaptations have been made to this architecture in the past few years for images~\cite{10.1145/3505244}, time series~\cite{10.24963/ijcai.2023/759}, and, naturally, tabular data~\cite{Huang2020TabTransformerTD, Somepalli2021SAINTIN, Zhang2023MixedTypeTD}. In the tabular domain, the purpose of the Transformer architecture is to capture meaningful relations between feature representations of the data via attention mechanisms.

\paragraph{Motivation}

Transformers are becoming a fundamental architectural block to model feature interactions in tabular data on different learning paradigms~\cite{Somepalli2021SAINTIN, Gorishniy2021RevisitingDL, Zhang2023MixedTypeTD}. Typically, they operate on a “raw” level of representation, i.e., at the data input level. In this work, we question its use to leverage abstract representations of a Neural Network, i.e., feature representations obtained by fully connected layers followed by nonlinearities. A prominent architecture for this study is the Variational Autoencoder (VAE), which encompasses three distinct kinds of representations: 1) an input representation that is fed into the recognition model (encoder); 2) a latent representation modeled via the statistics obtained by the recognition model; and 3) a reconstructed representation obtained by the generative model (decoder) (see Fig.~\ref{fig:base_architecture}). Fully Connected (FC) layers, which are affine maps, do not capture high-order dependencies between feature representations that may become important at deeper feature representations. The Transformer attention mechanism provides a suitable alternative to model these dependencies.

\paragraph{Our Contribution}

This work evaluates the impact of Transformers in a VAE for tabular data generation. We aim to answer the following question --- \textit{What's the impact of integrating Transformers into different components of a VAE?}
Starting with a VAE without Transformers, we study the effect of leveraging it over the aforementioned representations. In total, 6 variations are considered. This evaluation is performed by considering metrics that evaluate the statistical properties of the synthetic data with respect to the real data and through a Machine-Learning utility perspective. In addition, Center Kernel Alignment (CKA)~\cite{kornblith2019similarityneuralnetworkrepresentations} is used to compare similarities between feature representations on different components of the architectures. Our experiments are conducted using 57 datasets from the OpenML CC18 suite~\cite{oml-benchmarking-suites}.

From an evaluation perspective, our study reveals a trade-off between fidelity and diversity of the synthetic data with respect to real data. Specifically, we observe that as Transformers are added into the architecture, synthetic data tends to be less faithful to real data, while its diversity increases, with the biggest gain obtained when leveraging the latent and output representations with Transformers. From a representational perspective, our findings reveal that the input and output of Transformers at both the encoder and decoder tend to present a high degree of similarity. Moreover, at the decoder, the Transformer appears to act as the identity function due to little or no effects in representational changes in residual connections. We observe that the reason for this effect is due to layer normalization, which shifts and scales the initial representation such that it offers no representational changes.  

\paragraph{}

The paper is organized as follows: Section \ref{sec:related} reviews related work. Section \ref{sec:form} introduces the necessary formulations and provides an overview of the considered VAE models for the study. The experimental setup is outlined in Section \ref{sec:exp_setup}, followed by an analysis of the experimental results in Section \ref{sec:exp_results}. Section \ref{sec:cka} presents a detailed study on the similarities between feature representations at different levels of the architectures. Finally, conclusions are drawn in Section \ref{sec:conc}.

\section{Related Work}\label{sec:related}

Recent advances in tabular data generation draw upon a variety of deep generative approaches. Generative Adversarial Networks (GANs) such as CTGAN \cite{10.5555/3454287.3454946} and its successors CTAB-GAN \cite{zhao21} and CTAB-GAN+ \cite{zhao2023ctab} adapt GAN-based frameworks to handle continuous and categorical features. Meanwhile, Diffusion Models \cite{10.5555/3045118.3045358, 10.5555/3495724.3496298} have been specialized for tabular data through methods like TabDDPM \cite{10.5555/3618408.3619133}, CoDi \cite{Lee2023CoDiCC}, StaSy \cite{kim2023stasy}, TabSyn \cite{Zhang2023MixedTypeTD}, and TabDiff \cite{shi2025tabdiff}, each proposing distinct strategies to manage mixed data types and improve generative quality. Autoregressive models like GReaT~\cite{borisov2023language} and TabMT~\cite{gulati2023tabmtgeneratingtabulardata} have also been considered in the tabular data generation framework. Flow-Matching \cite{lipman2023flowmatchinggenerativemodeling} has led to novel gradient-boosting-based generative solutions \cite{jolicoeurmartineau2024generatingimputingtabulardata}. Finally, Variational Autoencoders (VAEs) \cite{Kingma2013AutoEncodingVB} adaptations include TVAE~\cite{10.5555/3454287.3454946}, introduced alongside CTGAN, and VAEM~\cite{10.5555/3495724.3496667}, which consists of a two-stage training — the first independently trains each feature using a VAE, while the second model's the inter-variable dependencies using the latent variables learned from the first stage. Another variation, GOGGLE~\cite{Liu2023GOGGLEGM}, was introduced as a generative model that approximates relational structure between variables via Graph Neural Networks, jointly training these relations with a VAE.

\paragraph{Transformers in Tabular Data Generation}

The Transformer architecture has been explored as a means to capture feature relationships in tabular data generation. For example, TabDiff employs a Transformer both in the backbone and at the denoiser output, whereas TabSyn leverages Transformers to model the statistics of the recognition model and the generative distribution of a VAE. Nevertheless, a thorough examination of how Transformers affect data quality remains an open question. In this work, we attempt to fill this gap.

\section{Formulation and Methods}\label{sec:form}

\begin{figure*}[!ht]
    \centering
    \includegraphics[width=\textwidth]{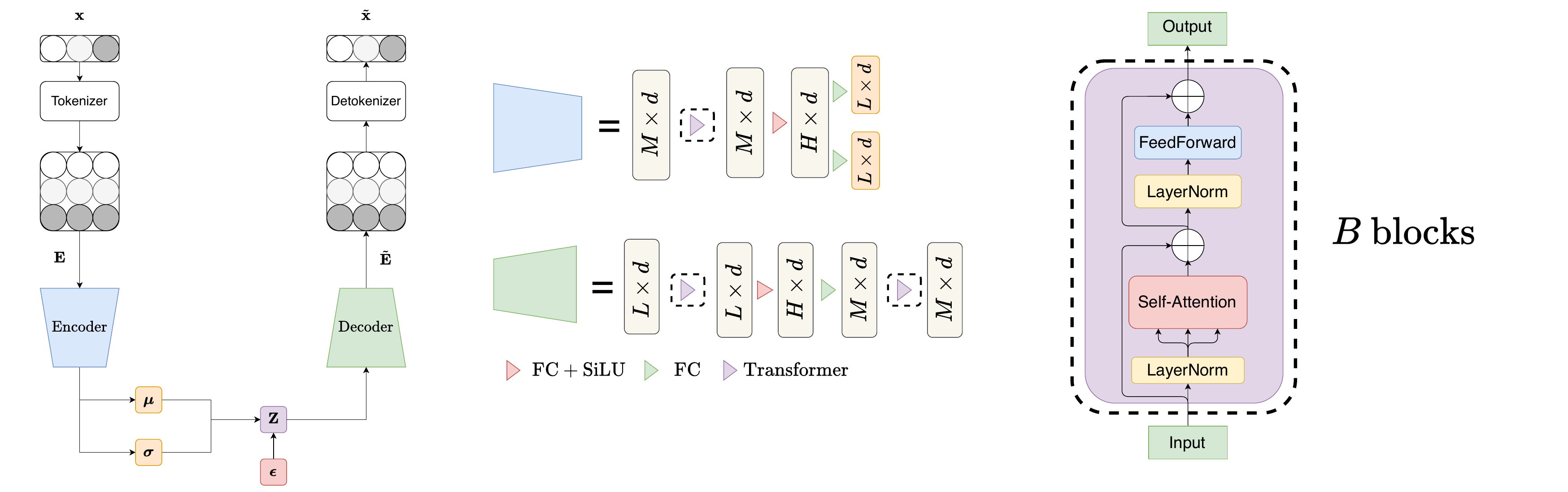}
    \caption{\textbf{Left}: Illustration of an embedding based VAE architecture. \textbf{Middle}: Encoder and Decoder mappings of the considered models. Each block denotes a feature map inside an encoder/decoder, with the respective input/output dimensions. Arrows denote operations performed on each feature representation, and the dashed rectangles describe the Transformer components detached in the considered methods. \textbf{Right}: The Transformer architecture implemented in this work.}
    \label{fig:base_architecture}
\end{figure*}

In the context of tabular data, datasets typically consist of mixed-type variables. In this paper, we focus on datasets that contain numerical and categorical features. A formulation of a dataset consisting of a mixture of these features follows.

Let $\mathcal{I} = \{\bx\}$ denote an instance of a dataset $\mathcal{D}$ with size $N$. We denote a data point $\bx$ to be represented  as a set of numerical $\bx^{(\text{num})} \in \mathbb{R}^{M_n}$ and categorical features $\bx^{(\text{cat})} \in \mathbb{R}^{M_c}$ as the following vector

\begin{equation}
    \bx = \left(x_1^{(\text{num})}, ..., x_{M_n}^{(\text{num})}, x_1^{(\text{cat})}, ..., x_{M_c}^{(\text{cat})}\right) \in \mathbb{R}^{M}~~,
\end{equation}

\noindent with $M=M_n+M_c$. Categorical variables $x_j^{(\text{cat})}$ are represented by a one-hot encoded vector, $\bx_j^{(\text{ohe})} \in \mathbb{N}^{|C_j|}$, where $C_j =\{1,..., |C_j|\}$ and $|C_j|$ denotes the number of categories of a given categorical feature $j$ such that in the end, each data point is represented as

\begin{equation}\label{eq:x_final}
    \bx = \left(x_1^{(\text{num})}, ..., x_{M_n}^{(\text{num})}, \bx_1^{(\text{ohe})}, ..., \bx_{M_c}^{(\text{ohe})}\right)~~ \in \mathbb{R}^{M'}~~,
\end{equation}

\noindent where $M' = M_n + \sum_{j=1}^{M_c} |C_j|$. 


\subsection{Embeddings Representation}\label{subsec:embvae}

As previously mentioned, one of the challenges in tabular data generation is to properly model its distribution due to the mixed-type nature of features. In this work, we tackle this problem by representing each feature as a continuous vector via tokenization~\cite{Gorishniy2021RevisitingDL, Zhang2023MixedTypeTD}.

\paragraph{Feature Tokenizer}\label{subsubsec:feature_token}

Let $\bx$ be the input of a neural network. A feature tokenizer takes as input a data point and projects it into a $(M \times d)$-dimensional space as:

\begin{equation}
\begin{split}
    \be_i^{(\text{num})} &= x_i^{(\text{num})} \textbf{w}_i^{(\text{num})} + \textbf{b}_i^{(\text{num})} \\
    \be_i^{(\text{cat})} &= \bx_i^{(\text{ohe})} \textbf{W}_i^{(\text{cat})} + \textbf{b}_i^{(\text{cat})}
\end{split} \label{eq:tokenizer}~~,
\end{equation}

\noindent where $\textbf{w}_i^{(\text{num})}, \textbf{b}_i^{(\text{num})}, \textbf{b}_i^{(\text{cat})} \in \mathbb{R}^{1 \times d}$ and $\textbf{W}_i^{(\text{cat})} \in \mathbb{R}^{|C_j| \times d}$. In other words, each numerical feature is projected into a vector space where each sample shares the same weights, while for categorical features, this tokenization acts as a lookup table, i.e., each category has its own set of learnable weights. In the end, $\bx$ is represented as the embedding matrix $\bE \in \mathbb{R}^{M \times d}$ by concatenating each $\be_i$ along the feature dimension, $\bE = \mathop{\bigoplus}_{i=1}^M \be_i$.

\paragraph{Feature Detokenizer}\label{subsubsec:feature_detoken}

Given a reconstructed embedding matrix $\Tilde{\bE} \in \mathbb{R}^{M \times d}$, the reconstructed representation $\Tilde{\bx}$ is obtained by projecting each embedding vector $\be_i$ back to the feature space as

\begin{equation}
\begin{split}
    \tilde{x}_i^{(\text{num})} &= \tilde{\be}_i^{(\text{num})} \tilde{\textbf{w}}_i^{(\text{num})} + \tilde{b}_i^{(\text{num})} \\ 
    \tilde{\bx}_i^{(\text{ohe})} &= \text{Softmax}\left(\tilde{\be}_i^{(\text{cat})} \tilde{\bW}_i^{(\text{cat})} + \tilde{\boldsymbol{b}}_i^{(\text{cat})}\right) \\
\end{split}~~,
\end{equation}

\noindent with $\tilde{b}_i^{(\text{num})} \in \mathbb{R}^{1 \times 1}$, $\tilde{\textbf{w}}_i^{(\text{num})} \in \mathbb{R}^{d \times 1},~ \tilde{\boldsymbol{b}}_i^{(\text{cat})} \in \mathbb{R}^{1 \times |C_j|}$ and $\tilde{\bW}_i^{(\text{cat})} \in \mathbb{R}^{d \times |C_j|}$. 
In the end, we concatenate every reconstructed feature

\begin{equation}
     \tilde{\bx} = \left(\tilde{x}_1^{(\text{num})}, ..., \tilde{x}_{M_n}^{(\text{num})}, \tilde{\bx}_1^{(\text{ohe})}, ..., \tilde{\bx}_{M_c}^{(\text{ohe})}\right)~~.
\end{equation}

\subsection{Self-Attention}\label{subsec:attn}

The main component of the Transformer architecture (cf. right image of Fig.~\ref{fig:base_architecture}) is the attention mechanism. In the context of tabular data, its purpose is to capture relationships between variables in the embedding space. In our work, this is accomplished via dot-product attention. These interactions go through a Softmax non-linearity to normalize the contribution of all features with respect to a given one. Letting $\textbf{Q} = \textbf{W}_{\textbf{Q}} \textbf{E}$, $\textbf{K} = \textbf{W}_{\textbf{K}} \textbf{E}$, $\textbf{V} = \textbf{W}_{\textbf{V}} \textbf{E}$ be a set of query, key and values, respectively, the attention mechanism outputs the weighted sum of values $\textbf{V}$

\begin{equation}
    \text{Attention}(\textbf{Q}, \textbf{K}, \textbf{V}) = \text{Softmax}\left(\frac{\textbf{Q}\textbf{K}^T}{d_k}\right)\textbf{V}~~,
\end{equation}

\noindent where $d_k$ is the embedding dimensionality of $\textbf{K}$.

\subsection{Models}\label{sec:methods}

The models under study in this work are summarized in Table~\ref{tab:models} and follow the scheme illustrated in the first image of Fig.~\ref {fig:base_architecture}. An initial implementation that leverages embeddings at the input and output, but without Transformers, is considered, and we call it VAE. The following models consider Transformers acting at the backbone of the encoder, the latent space, and the decoder head. The different parts of the network where these Transformers are included are explicitly shown in the middle image of Fig.~\ref{fig:base_architecture}. We detail a forward pass of the architecture that leverages Transformers over all the considered (ELD-VAE) in the Supplemental Material. The remainder of the architectures share the same encoder and decoder components, except where the Transformer acts. For a review of the theory behind Variational Autoencoders, we recommend the readers to~\cite{10.1561/2200000056}.

\begin{table}[ht]
    \centering
    \scriptsize
    \caption{Architectures under study in this work.}
    \begin{tabular}{l|*{3}{c}}
        \toprule
        \textbf{Model (Abbreviation)} & \multicolumn{3}{c}{\textbf{Transformer on:}} \\
        \midrule
        & Enc(oder) & Lat(ent) & Dec(oder)  \\
        VAE (VAE) & \xmark & \xmark & \xmark \\
        Enc-VAE (E-VAE) & \cmark & \xmark & \xmark \\
        Enc-Lat-VAE (EL-VAE) & \cmark & \cmark & \xmark \\
        Enc-Lat-Dec-VAE (ELD-VAE) & \cmark & \cmark & \cmark \\
        Lat-Dec-VAE (LD-VAE) & \xmark & \cmark & \cmark \\
        Dec-VAE (D-VAE) & \xmark & \xmark & \cmark \\
        \bottomrule
    \end{tabular}
    \label{tab:models}
\end{table}

\section{Experimental Setup}\label{sec:exp_setup}

\subsection{Datasets}\label{subsec:datasets}

We use the OpenML CC18 suite~\cite{oml-benchmarking-suites} as a benchmark to evaluate the methods presented in this paper. It is composed of 72 datasets used for classification tasks. From this benchmark, we select 57 datasets that encompass samples and feature dimensions in the range between $N \in [500, 96320]$ and $M \in [4, 240]$, respectively. For all datasets, the train and test splits provided by the OpenML CC18 suite are used, and finally, we extract 15\% of the training set, which serves as our validation set. We observed training instabilities in three of the considered datasets, which we detail in the Supplemental Material.

For all datasets, the following pre-processing is applied: 1) we begin by dropping features that only contain missing values, and numerical or categorical features with 0 variance or only one category, respectively; 2) numerical and categorical columns with missing values are replaced with the mean and mode, respectively; 3) numerical variables are encoded using a quantile Transformer with a Gaussian distribution~\footnote{\url{https://scikit-learn.org/stable/modules/generated/sklearn.preprocessing.QuantileTransformer.html}} based on previous works~\cite{Zhang2023MixedTypeTD, shi2025tabdiff}, while categorical variables are encoded using one-hot encoding.

\subsection{Training and Sampling Details}\label{subsec:train_details}

Each model is trained under 500 epochs with the Adam optimizer \cite{kingma2017adammethodstochasticoptimization}, using a weight decay of $0.9$ and a learning rate of $1 \times 10^{-3}$. The batch size is determined by following simple rules, given the validation set. The tokenizer, detokenizer, hidden, and latent layers weights are initialized with the same values for all models, and if two architectures share the same Transformer location, so the Transformer.

Regarding model hyperparameters, we keep them constant over all datasets and models. Each Transformer is defined with one head, four blocks~\footnote{In the Supplemental Material, we study the effect of lowering the number of blocks in terms of high-density estimation metrics.}, a hidden dimension of 128, and without dropout. By recommendation~\cite{Gorishniy2021RevisitingDL}, we also use its pre-norm variation~\cite{Wang2019LearningDT}. An embedding dimension of $d=4$ is considered following previous works~\cite{Zhang2023MixedTypeTD}. The fully connected layers of the encoder and decoder have a hidden dimension of $H=128$, while the latent dimension is $L=64$, unless stated otherwise.

After training, we sample over the latent as $\bz \sim \mathcal{N}(\boldsymbol{0}, \textbf{I})$. In our experiments, the number of synthetic samples is the same size as the training data.

\subsection{Evaluation Metrics}\label{subsec:eval_metrics}

The synthetic data produced by the generative models under study are evaluated using several metrics found in the literature. We divide the considered metrics into three groups: 1)~\textit{Low-Density Estimation}, where 1-way marginals and pairwise-correlations measurements are considered to estimate differences between feature distributions; 2)~\textit{High-Density Estimation}, which compares the joint probability distributions of synthetic and real data; 3)~\textit{Machine Learning-Efficiency}, aiming to determine the usefulness of synthetic data in downstream tasks such as classification. 

Note that all metrics are defined on a domain between [0, 1], where the higher the value, the better the model performance is.

\subsubsection{Low-Density Estimation}

Under this class of metrics, we consider 1-way marginals and pairwise correlations.

\paragraph{1-Way Marginals} The first metric measures how similar the (independent) feature distributions between real and synthetic data are. The Kolmogorov-Smirnov statistic~\cite{Hodges1958TheSP} is computed for numerical columns, under the null hypothesis that real and synthetic data are drawn from the same probability distribution, while the Total Variation Distance~\cite{LevinPeresWilmer2006} is applied for categorical ones. In the end, we average the similarities obtained from each feature.

\paragraph{Pairwise-Correlations} Pairwise-correlations measure the dependency between two features in a dataset. Given two columns $(m_1, m_2)$ of both $(\bx, \tilde{\bx})$, if they are both numerical, we determine Pearson's Correlation~\cite{doi:10.1098/rspl.1895.0041}; if they are both categorical, the Contingency Similarity; finally if they are of different types, the numerical column is partitioned into bins, and afterwards, the Contingency Similarity is applied. The score between the correlations $(\rho, \tilde{\rho})$ obtained for each type of data is then determined as

\begin{equation}
    score = 1 - \frac{|\rho_{m_1, m_2} - \tilde{\rho}_{m_1, m_2}|}{2}~~.
\end{equation}

Finally, we average all the scores obtained for each pairwise correlation. For these two first metrics, we use the implementations provided by the \texttt{sdmetrics}~\cite{sdmetrics} Python package. In our experiments, we dub these metrics as low-density, as they only capture unidimensional statistics and pairwise relationships between variables.

\setlength{\tabcolsep}{5pt} 
\begin{table*}[!ht]
    \centering
    \scriptsize 
    \caption{Average results obtained for the studied models. For a given metric, a performer with the highest average score is highlighted in \textbf{bold}, while with the lowest average rank \underline{underlined}. ($^{**}$) implies that the given model results are statistically significant with a $p$-value of 0.001 using Wilcoxon's Signed Rank Test with respect to the best performer.}
\begin{tabular}{l|cc|cc|cc|cc|cc|cc}
& \multicolumn{4}{c|}{Low-Density} & \multicolumn{4}{c|}{High-Density} & \multicolumn{4}{c}{ML-Efficiency} \\
\midrule
& \multicolumn{2}{c|}{Marginals~($\uparrow$)} & \multicolumn{2}{c|}{Pairwise-Correlations~($\uparrow$)} & \multicolumn{2}{c|}{$\alpha$-Precision~($\uparrow$)} & \multicolumn{2}{c|}{$\beta$-Recall~($\uparrow$)} & \multicolumn{2}{c|}{Utility~($\uparrow$)} & \multicolumn{2}{c}{ML-Fidelity~($\uparrow$)} \\
\midrule
& Score & Rank & Score & Rank & Score & Rank & Score & Rank & Score & Rank & Score & Rank \\
Model &  &  &  &  &  &  &  &  &  &  &  &  \\
\midrule
VAE & 0.913 $\pm$ \tiny{0.048} & \underline{3.20} & 0.924 $\pm$ \tiny{0.051} & 3.30 & 0.801 $\pm$ \tiny{0.177} & \underline{2.60} & 0.317 $\pm$ \tiny{0.247}$^{**}$ & 4.80 & 0.770 $\pm$ \tiny{0.174} & 3.90 & 0.805 $\pm$ \tiny{0.169} & 4.00 \\
E-VAE & 0.912 $\pm$ \tiny{0.047} & 3.30 & 0.921 $\pm$ \tiny{0.052} & 3.90 & \textbf{0.803 $\pm$ \tiny{0.168}} & \underline{2.60} & 0.314 $\pm$ \tiny{0.248}$^{**}$ & 5.10 & 0.774 $\pm$ \tiny{0.169} & 3.60 & 0.805 $\pm$ \tiny{0.167} & 3.70 \\
EL-VAE & 0.910 $\pm$ \tiny{0.050} & 4.10 & 0.924 $\pm$ \tiny{0.051} & 3.60 & 0.769 $\pm$ \tiny{0.179} & 3.50 & 0.361 $\pm$ \tiny{0.236}$^{**}$ & 3.20 & 0.777 $\pm$ \tiny{0.161} & 3.20 & 0.811 $\pm$ \tiny{0.158} & 3.30 \\
ELD-VAE & 0.916 $\pm$ \tiny{0.038} & 3.60 & 0.926 $\pm$ \tiny{0.048} & 3.40 & 0.749 $\pm$ \tiny{0.189}$^{**}$ & 4.20 & 0.388 $\pm$ \tiny{0.244} & 2.40 & 0.776 $\pm$ \tiny{0.174} & \underline{3.00} & 0.815 $\pm$ \tiny{0.157} & \underline{3.20} \\
LD-VAE & \textbf{0.917 $\pm$ \tiny{0.039}} & 3.30 & \textbf{0.928 $\pm$ \tiny{0.048}} & \underline{3.10} & 0.752 $\pm$ \tiny{0.194}$^{**}$ & 3.90 & \textbf{0.392 $\pm$ \tiny{0.243}} & \underline{2.10} & \textbf{0.778 $\pm$ \tiny{0.161}} & 3.30 & \textbf{0.817 $\pm$ \tiny{0.152}} & \underline{3.20} \\
D-VAE & 0.903 $\pm$ \tiny{0.081} & 3.50 & 0.917 $\pm$ \tiny{0.075} & 3.70 & 0.734 $\pm$ \tiny{0.226}$^{**}$ & 4.10 & 0.359 $\pm$ \tiny{0.255}$^{**}$ & 3.40 & 0.749 $\pm$ \tiny{0.173} & 4.00 & 0.791 $\pm$ \tiny{0.160} & 3.70 \\
\bottomrule
\end{tabular}
\label{tab:tab_main_res}
\end{table*}
\setlength{\tabcolsep}{6pt} 

\subsubsection{High-Density Estimation} 

These metrics compare the joint distribution of real and synthetic data. We use the work from~\cite{Alaa2021HowFI}, which introduces the notion of $\alpha$-Precision and $\beta$-Recall. Generally speaking, $\alpha$-Precision and $\beta$-Recall characterize the fidelity and diversity of the generated data to the real one, respectively. While $\alpha$-Precision is computed by determining the probability that a generated sample resides in the support of the real-data distribution, $\beta$-Recall is computed by determining the probability that a real sample resides in the support of the synthetic data distribution. A synthetic data point $\tilde{x}_n$ is said to reside inside the $\alpha$-support of the real distribution if 

\begin{equation}
    f_\alpha(\tilde{x}_n) = \boldsymbol{1}\{\tilde{x}_n \in \textbf{B}(c_r, r_\alpha)\}~~,
\end{equation}

\noindent where $\textbf{B}(c_r, r_\alpha)$ is a non-parametric estimator of the support of real data, assumed to be a ball of radius $r_{\alpha} = Q_\alpha\{||x_n - c_r||:1 \leq n \leq N\}$ and center $c_r = \sum_{i=1}^M x_i$. $Q_\alpha$ is the $\alpha$-quantile function. On the other hand, a real data point $x_n$ is said to reside inside the $\beta$-support of the synthetic distribution if

\begin{equation}
    f_\beta(x_n) = \boldsymbol{1} \{\tilde{x}_{n^*}^{\beta} \in \textbf{B}(x_n, \text{NND}_k(x_n))\}~~,
\end{equation}

\noindent where $\tilde{x}_{n^*}^{\beta}$ is the synthetic sample in the ball closest to $x_n$, and $\text{NND}_k(x_n)$ the $k$-nearest neighbor in $\mathcal{D}$. After averaging $f_\alpha(\tilde{x}_n)$ and $f_\beta(x_n)$ for all data points, these quantities are integrated over all ($\alpha, \beta$)-quantiles. We use the implementation of these metrics provided by the \texttt{synthcity} package~\cite{doi.org/10.48550/arxiv.2301.07573}.

\subsubsection{Machine Learning-Efficiency}

Regarding Machine Learning-Efficiency (ML-Efficiency), we are interested in both Utility and ML-Fidelity. The classifier taken into consideration is XGBoost~\cite{10.1145/2939672.2939785}. Hyperparameters are kept constant, with a number of 500 boosters and a learning rate of $1 \times 10^{-2}$. After training a model, a real test set is evaluated using two models --- one trained over real data, $\mathcal{M}_{\text{real}}$, and another trained over synthetic data, $\mathcal{M}_{\text{syn}}$. We denote predictions obtained from $\mathcal{M}_{\text{real}}$ and $\mathcal{M}_{\text{syn}}$ as $\hat{y}^{(\text{real})}$, $\hat{y}^{(\text{syn})}$, respectively.

\paragraph{Utility} By utility, we ask how well a model performs when trained over a synthetic dataset $\Dsyn$ and evaluated under a holdout set from the real dataset $\bx^{(\text{test})}$. We adopt the Train on Synthetic, Test on Real (TSTR) (e.g.~\cite{Esteban2017RealvaluedT}) methodology. Here, predictions are evaluated using accuracy.

\paragraph{ML-Fidelity} By ML-Fidelity, we ask how similar the predictions ($\hat{y}^{(\text{real})}$, $\hat{y}^{(\text{syn})}$) are. This metric is also measured in terms of accuracy, i.e.

\begin{equation}
    \text{ML-Fidelity} = \frac{1}{\left|\mathcal{D}^{(\text{test})}\right|} \sum_{i=1}^{\left|\mathcal{D}^{(\text{test})}\right|} \boldsymbol{1}\left(\hat{y}_i^{(\text{real})} = \hat{y}_i^{(\text{syn})}\right)~~.
\end{equation}

\subsection{Implementation Details}

Models are implemented with Python's programming language using JAX ecosystem~\cite{jax2018github} and trained on a Linux Machine with 16GB of RAM and an NVIDIA RTX 2000 GPU\footnote{We will publicly release the code once the paper is reviewed.}.

\section{Experimental Results}\label{sec:exp_results}

In this section, we perform a systematic evaluation to understand the impact of Transformers on different components of a VAE architecture. We begin by reporting results based on average results and ranks in Section~\ref{subsec:overview}. In Section~\ref{subsec:fid-div-tradeofff}, the observed trade-off between fidelity and diversity is detailed.

\subsection{General overview}\label{subsec:overview}

We begin by reporting top performers based on average score and ranking for each metric based on the results presented in Table~\ref{tab:tab_main_res}. Models with the highest average score are highlighted in \textbf{bold}, while those with the highest average rank are \underline{underlined}. In addition, for given metrics, models with double asterisks ($^{**}$) imply that their results are statistically significant to the top performer using the Wilcoxon's Signed Rank Test~\cite{c4091bd3-d888-3152-8886-c284bf66a93a}.

Overall, LD-VAE consistently excels across the considered metrics, attaining the highest average score except for $\alpha$-Precision, where the E-VAE outperforms it. Although LD-VAE achieves the best mean scores on most metrics, its performance on low-density estimation and ML-Efficiency metrics does not present statistically significant differences compared to other variations. However, for high-density metrics, LD-VAE results on $\beta$-Recall are statistically superior to those of all other models except ELD-VAE. In contrast, for $\alpha$-Precision, the  E-VAE achieves statistically significant results compared to models that leverage Transformers in the decoder (D-VAE, LD-VAE, ELD-VAE).

Ranking-wise, there is no single clear winner. In low-density estimation metrics, despite LD-VAE having the highest score, the baseline VAE attains the best rank for 1-way marginals. Meanwhile, LD-VAE claims the top rank for Pairwise-Correlations. In high-density metrics, both top performers share the lowest rank; notably, the baseline VAE ties with the encoder-only variant on $\alpha$-Precision. Regarding ML-Efficiency, ELD-VAE holds the best rank in Utility, while competing with LD-VAE in terms of ML-Fidelity.

In sum, incorporating Transformers into different architecture components generally yields no substantial improvements for feature distribution modeling or machine-learning utility. The most notable effect appears in the high-density estimation metrics, where the baseline VAE and E-VAE synthesize data with higher fidelity, while models leveraging Transformers in the latent and decoder spaces produce more diverse samples. Section \ref{subsec:fid-div-tradeofff} delves deeper into this fidelity–diversity trade-off, and corresponding analyses for low-density estimation and ML-Efficiency are provided in the Supplemental Material.

\subsection{Fidelity-Diversity trade-off}\label{subsec:fid-div-tradeofff}

\begin{figure}
    \centering
    \includegraphics[width=0.23\textwidth]{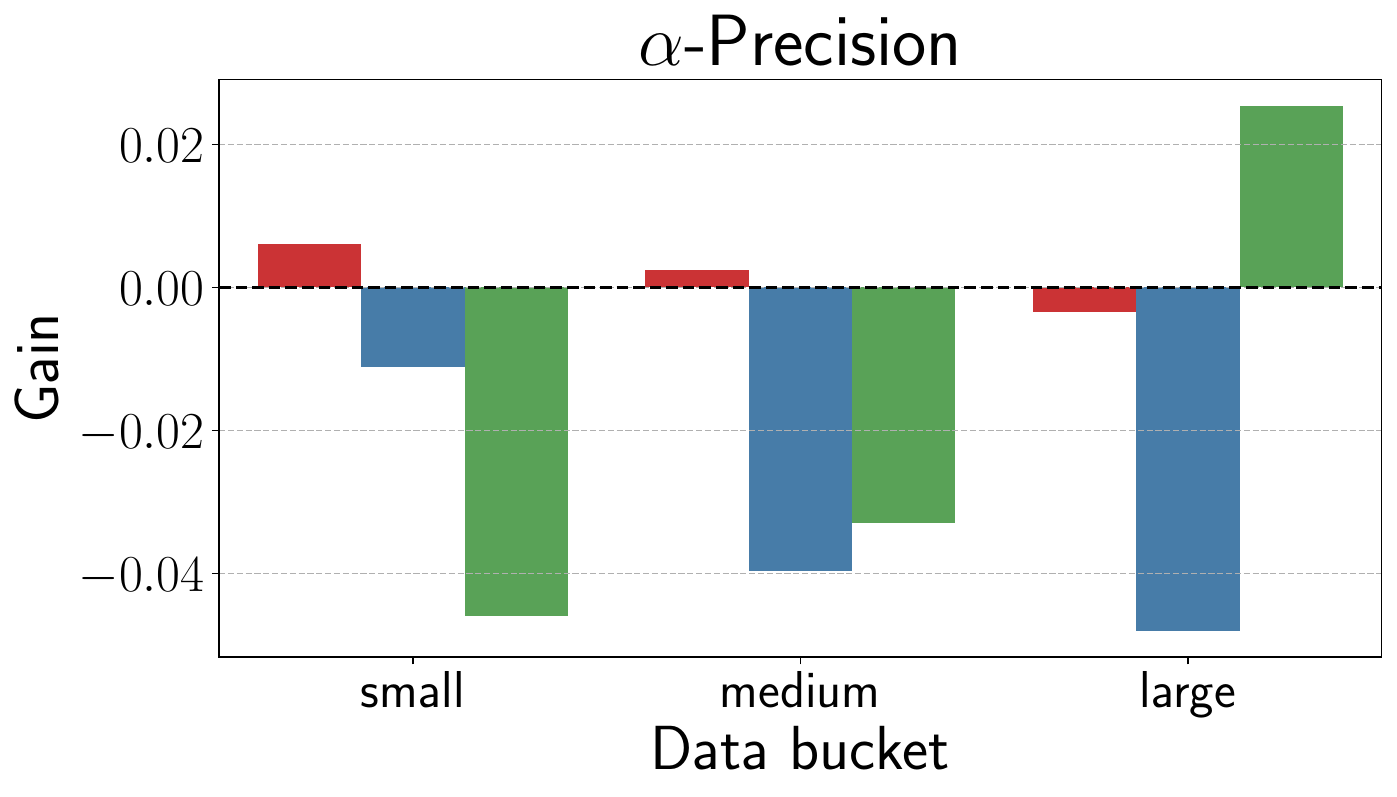}
    \includegraphics[width=0.23\textwidth]{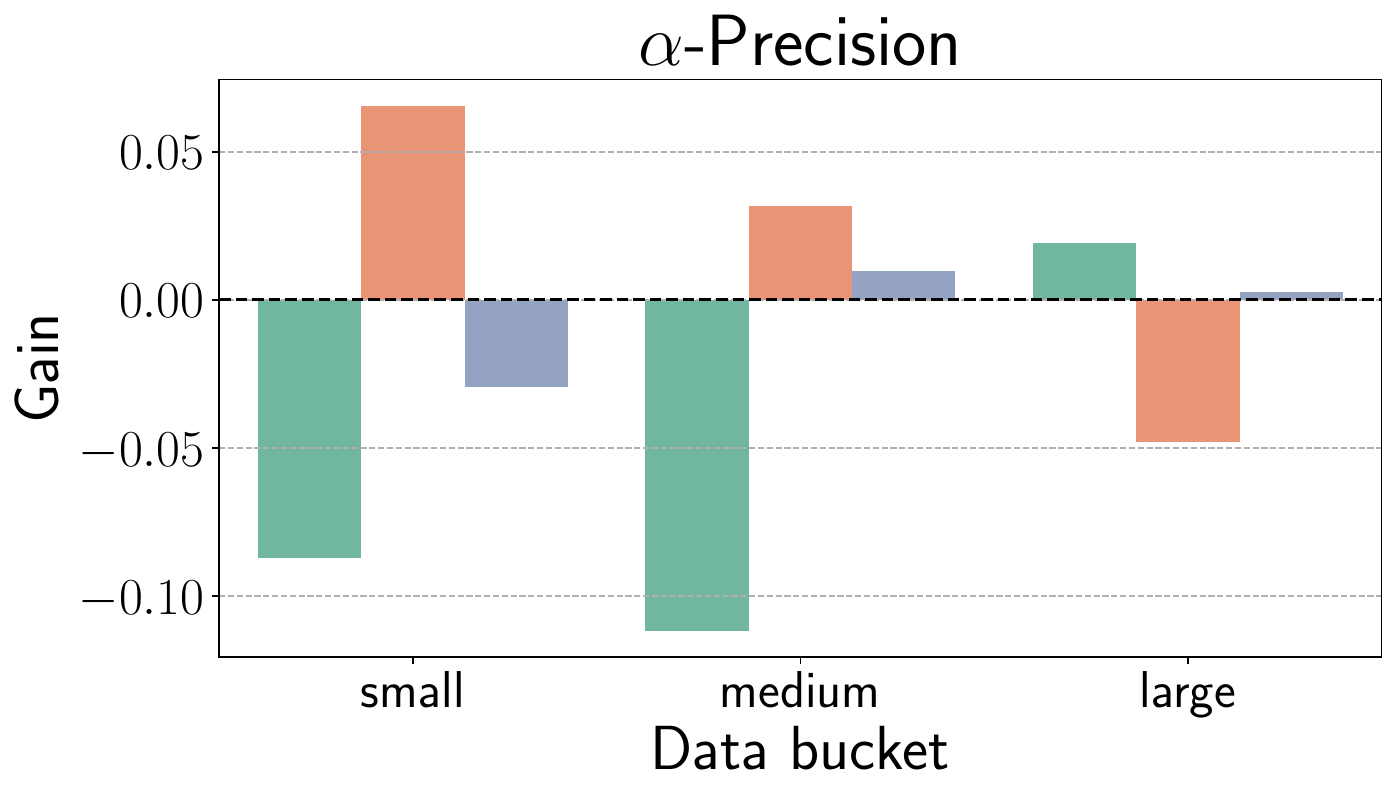}\\
    \includegraphics[width=0.23\textwidth]{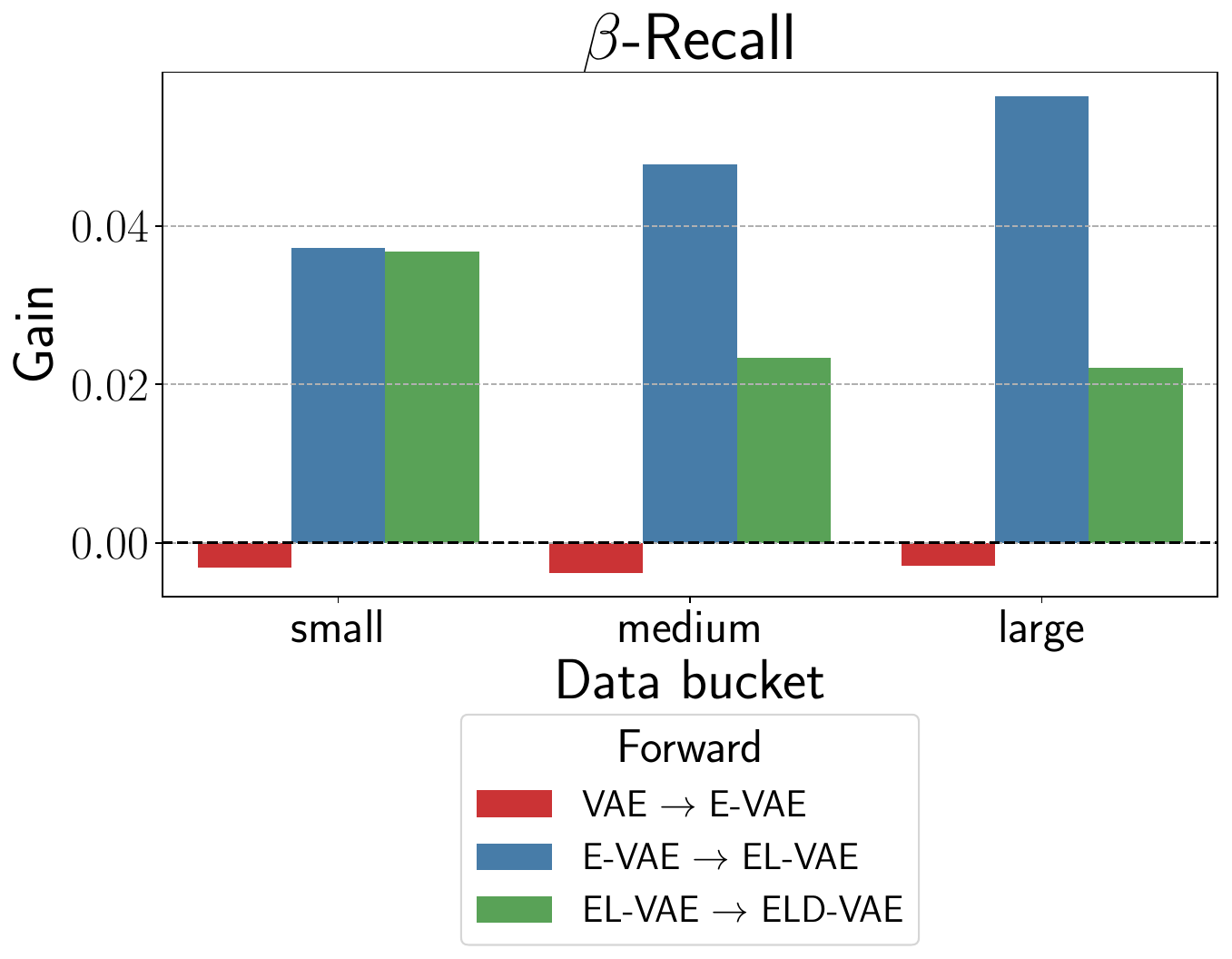}
    \includegraphics[width=0.23\textwidth]{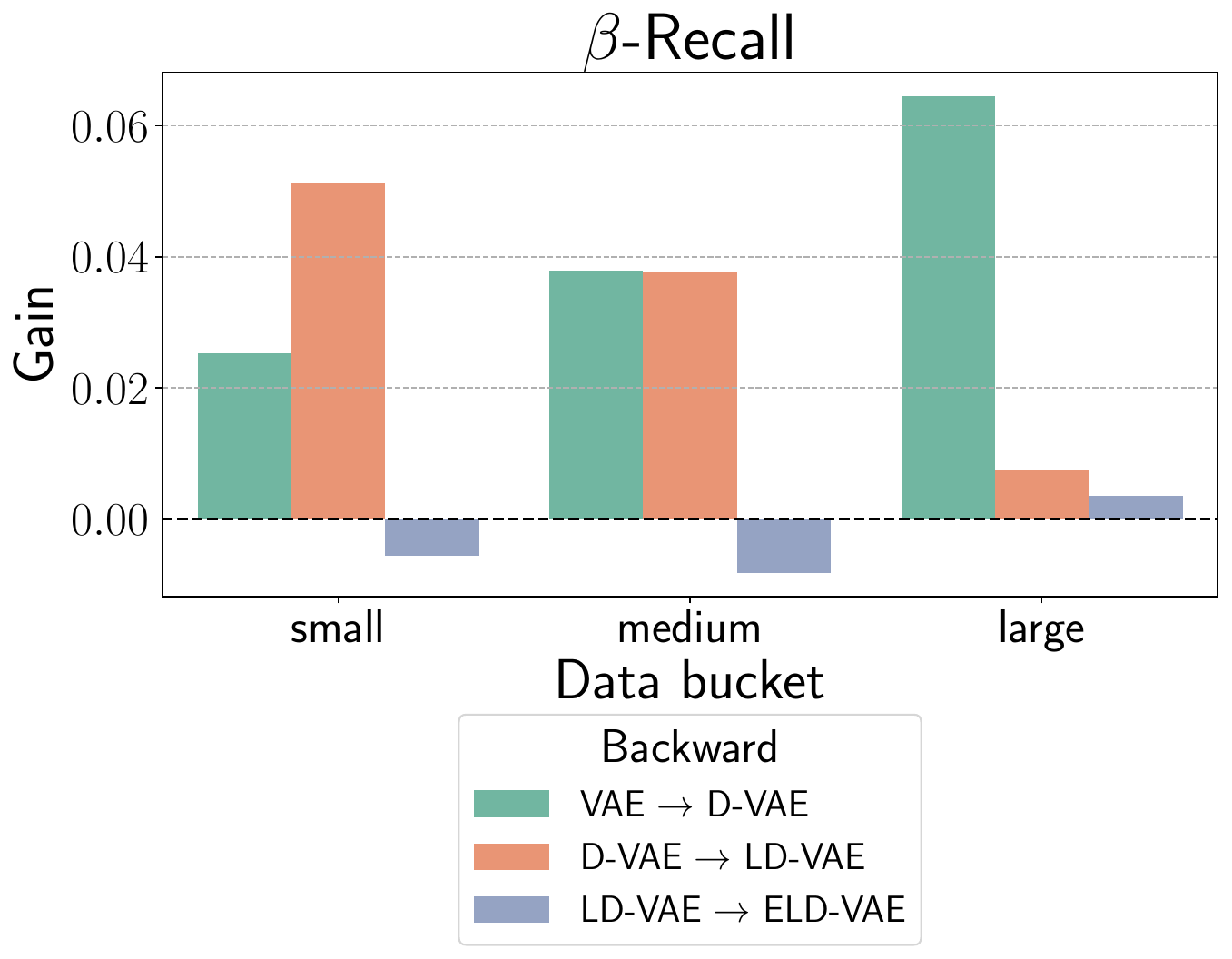}
    \caption{Aggregated gains in performance as Transformers are added to the studied components for $\alpha$-Precision (top-plots) and $\beta$-Recall (bottom-plots), for the Forward (left-plots) and Backward (right-plots) sequences.}
    \label{fig:gains}
\end{figure}

We analyze changes in performance as Transformers are added to the network. Two sequences of models are considered to draw conclusions --- one that begins by appending a Transformer into the encoder, then to the latent space, and finally to the decoder. We dub this sequence \textbf{Forward}, which follows VAE $\to$ E-VAE $\to$ EL-VAE $\to$ ELD-VAE; the second sequence begins by adding a Transformer to the decoder, then to the latent space, and finally to the encoder. We call this sequence \textbf{Backward} and follows VAE $\to$ D-VAE $\to$ LD-VAE $\to$ ELD-VAE. Note that the models considered at the beginning and end of the sequences are the same. We continue to use Table~\ref{tab:tab_main_res}, supported by Fig.~\ref{fig:gains}, depicting differences in performance as Transformers are added to the network as a function of dataset size buckets, by considering buckets from small to large with the following intervals: small $\in$ [500, 1000), medium $\in$ [1000, 5000) and large $\in$ [5000, 96230].

\paragraph{Forward Sequence}

Table~\ref{tab:tab_main_res} reveals that, with the exception of E-VAE, adding Transformers into various components of the architecture generally results in synthetic data that is less faithful to real data. In the transition from E-VAE $\to$ EL-VAE, it becomes clear that as the dataset size increases, the faithfulness of the synthetic data decreases (cf. top-right plot of Fig.~\ref{fig:gains}) by considering a Transformer to leverage the latent representation of the architecture. Conversely, when transitioning from EL-VAE to ELD-VAE, the trend reverses: while EL-VAE faithfulness diminishes, particularly in larger datasets, its performance remains superior for small to mid-size datasets until ELD-VAE eventually overtakes it on larger ones.

Regarding $\beta$-Recall, E-VAE performance remains approximately on par with that of the base VAE for all dataset buckets. On the other hand, an average gain of $4.7\%$ is obtained when considering a Transformer over the latent space. In addition, we observe a consistent increase in diversity as the data bucket size increases. An additional benefit of 2.7\% is achieved for applying a Transformer in the decoder (EL-VAE $\to$ ELD-VAE). The bottom-left plot of Fig.~\ref{fig:gains} further indicates that these gains are more significant on small-size datasets while remaining approximately the same for medium and large dataset buckets.

\paragraph{Backward Sequence}

In the backward sequence, a considerable drop in performance of 6.7\% in $\alpha$-Precision is observed by considering a Transformer to leverage the reconstructed representation of the data (D-VAE). As shown in the top-right plot of Fig.~\ref{fig:gains}, this decrease is consistent across all dataset sizes except for the larger ones, where a 2\% improvement emerges. A gain in fidelity of 6\% and 3.1\% is observed for small and mid-size datasets when leveraging a Transformer over the latent space, while a significant loss of 4.8\% is observed for large-size datasets. The transition from LD-VAE to ELD-VAE tells us that, for small datasets, leveraging the input representation with a Transformer deteriorates the faithfulness of the synthetic data, while for larger datasets, there is little to no gain. To conclude, in terms of $\beta$-Recall, notable gains are observed from leveraging Transformers over the decoder (VAE $\to$ D-VAE) and latent space (D-VAE $\to$ LD-VAE).

\paragraph{Discussion}

The evaluation presented in the previous section indicates a trade-off between high-density estimation metrics. In general, adding Transformers into a VAE posits a lower fidelity of the synthetic data w.r.t. the real one but a higher diversification. Comparing the base implementation of a VAE with the Transformed-based architectures, this trade-off is observed for all models except for E-VAE, which leverages the initial representation of the data at a lower level of abstraction (i.e., at the input level). We argue that E-VAE does not provide significant differences to its base implementation in terms of high-density estimation metrics since, after the model is trained, the encoder is ``detached" from the model, and only the generator (decoder) is used to synthesize samples. In the Supplemental Material, an additional discussion of this trade-off is considered by evaluating the reconstructed data obtained by a forward pass on the test set.

As deeper representations of the network are considered, self-attention has more flexibility in modeling feature interactions, as these representations are less constrained by the original representation of the data. From an evaluation perspective, we've shown in the previous section that this additional flexibility tends to produce more diverse and less faithful data. It is known that VAE produces less faithful (blurry) data~\cite{10.5555/3454287.3455618} in the image modality due to the regularization term in the loss function that tries to approximate the distribution of the recognition model with the Gaussian prior.
Here, we observe that considering Transformers to leverage latent and reconstructed representations further impacts this faithfulness, albeit observing a considerable gain in diversity.

From a Machine-Learning standpoint, Table \ref{tab:tab_main_res} shows that placing Transformers at various points in the network does not yield statistically significant gains. This outcome likely arises from spurious correlations introduced between the independent and dependent features, which, despite increasing diversity, do not translate into better decision-boundary estimation.

\section{Representations Similarities}\label{sec:cka}

\begin{figure}
    \centering
    \includegraphics[width=\columnwidth]{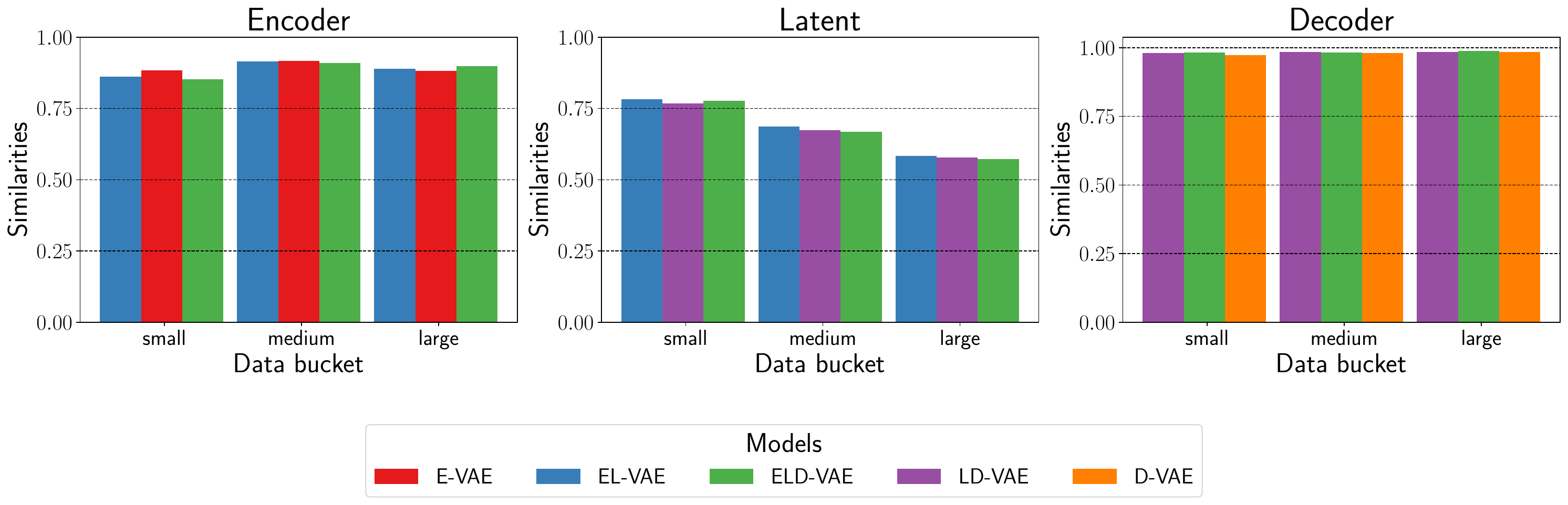}
    \caption{Aggregated similarities for the considered models for each data bucket. Each bar denotes the average similarity measure over all datasets, obtained between the input and output of a Transformer in the corresponding architecture component for the considered models.}
    \label{fig:similarities}
\end{figure}

We also investigate similarities between representations inside the considered architectures, aiming to understand the behavior of Transformers that act at the input, latent, and reconstructed representations. We use Center Kernel Alignment (CKA), which provides a similarity measure via dot-product between feature representations while preserving orthogonal transformations and isotropic scaling ~\cite{kornblith2019similarityneuralnetworkrepresentations}. Let $\bE_1 \in \mathbb{R}^{b\times l_1}$ and $\bE_2 \in \mathbb{R}^{b \times l_2}$, be two feature maps, where $b$ denotes the size of a batch, and $l_{1,2}$ the dimensionality of a given layer. The (linear) CKA similarity is defined as

\begin{equation}
    \text{CKA} = \frac{||\bE_2^T\bE_1||_F^2}{||\bE_1^T\bE_1||_F ||\bE_2^T\bE_2||_F}~~,
\end{equation}

\noindent where $||.||_F$ denotes the Frobenius norm. This measure is bounded between [0,1], where 1 denotes that the considered representations are similar.

\begin{figure}
    \centering
    \includegraphics[width=\columnwidth]{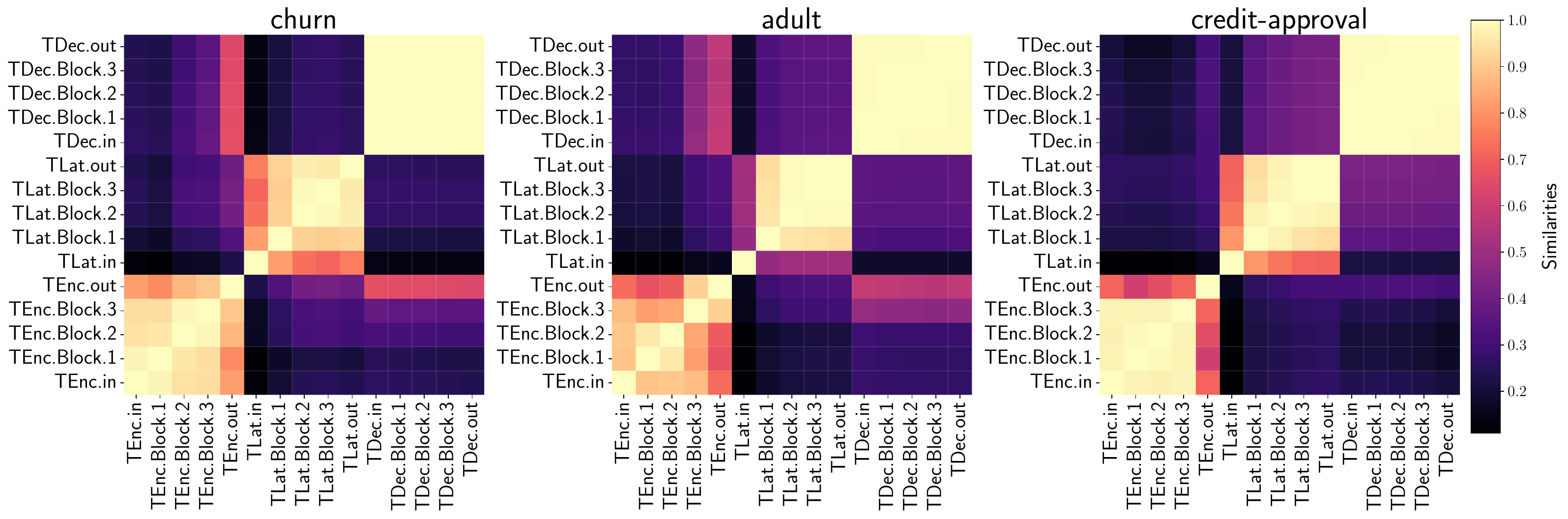}
    \caption{Similarities between block representations of a Transformer for the \texttt{churn}, \texttt{adult} and \texttt{credit-approval} datasets. Each cell of the heatmap denotes the similarity between two feature representations. Higher similarities have a lighter color. Depending on where a Transformer acts, we follow the naming convention T(Enc, Lat, Dec) to denote a given Transformer, while (in, out, block.i) to denote input, output, and internal block layer representations.}
    \label{fig:datasets_similarities}
\end{figure}

We begin by considering similarities between the input and output representations of Transformers that are leveraged on the studied representations (cf. the Transformer architecture in Fig.~\ref{fig:base_architecture}). Note that these representations are a tensor $\mathbf{E} \in \mathbb{R}^{b \times F \times d}$. In the following study, we flatten these representations into a matrix of shape $\mathbf{E} \in \mathbb{R}^{b \times Fd}$. The CKA similarity is always evaluated over data points of the test set of a given dataset in a fully trained model.

\paragraph{Aggregated Similarities}\label{subsec:agg_sims}
We begin by observing a high average similarity between representations before and after the encoder and decoder Transformers on all the considered networks, independently of the considered data bucket (cf. Fig.~\ref{fig:similarities}), hinting that, after a model is trained, and on average, little to no variation in the direction of these representations w.r.t. to its input are observed. Conversely, we observe a lower averaged similarity measure at the latent Transformer, likely due to the stochastic effects introduced by the reparameterization trick.

\begin{figure*}
    \centering
    \includegraphics[width=\textwidth]{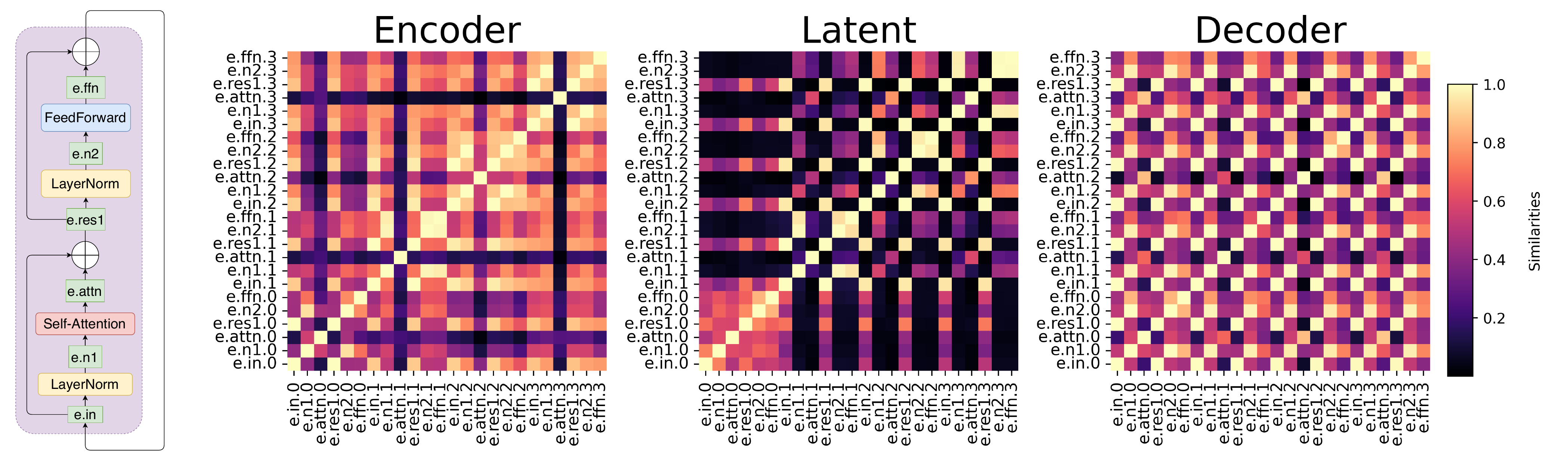}
    \caption{Heatmap similarities between representations on the considered Transformers for the ELD-VAE model, for the \texttt{adult} dataset. Each cell denotes the similarity between two feature representations inside a Transformer. The higher the similarity, the lighter the color of the cell. The Transformer on the left details the representations we extract to measure similarities presented in the heatmaps.}
    \label{fig:inner_transformer}
\end{figure*}

\paragraph{Internal Similarities} 

Following, we investigate the similarities between the input and output of each Transformer block of ELD-VAE on three different datasets. These similarities are presented in Fig.~\ref{fig:datasets_similarities}. We observe high similarities between consecutive Transformer block representations in each component. Notably, representations over the Transformer that act at the decoder present high similarity regardless of their depth, indicating little to no variation in the representation before and after its application, as previously discussed. Regarding Transformers that act at the input and latent representations, we observe a higher dissimilarity between the input and output representations, especially at the latent space, indicating representational changes in this region. We questioned whether this effect is due to possible over-parameterization of the network by considering different numbers of Transformer blocks in the considered components. We observed that even when one block is considered, a high similarity between representations is obtained. We explore this in more detail in the Supplemental Material.

Also, representations between different Transformer components (e.g., from Latent to Decoder) tend to be highly dissimilar, possibly due to leveraging these representations by the following pointwise non-linearities operating in both the encoder and decoder.

We further analyze how the internal representation similarities evolve within each Transformer block of the ELD-VAE architecture when applied to the \texttt{adult} dataset. Fig.~\ref{fig:inner_transformer} illustrates these changes, complemented with a Transformer block depicting the representations that are compared. It is noteworthy that although operations like layer normalization (\texttt{e.n1}) and self-attention (\texttt{e.attn}) initially exhibit lower similarity compared to the original representation (\texttt{e.in}), a significant increase in similarity is observed after applying the residual connection (\texttt{e.res}). While at the Transformers placed in the encoder and latent components, this effect is observed with a higher similarity inside each block; in the decoder, we observe this consecutively after each residual connection, across all blocks.

Recalling that the residual connection is the addition between the initial representation and the self-attention output, this suggests that the combination of layer normalization and self-attention does not alter the direction of the original representation. Instead, these components appear to function primarily as a scaling factor for the initial representation within each block. To validate this observation, we consider the residual connection expressed by $\hat{\bE} = \bE + f(\bE)$, where \(\hat{\bE}\) corresponds to \texttt{e.res}, \(f(\bE)\) to \texttt{e.attn}, and \(\bE\) to \texttt{e.in}. If this residual connection merely scales the initial representation, then \(\hat{\bE} = \sigma \bE\). Analyzing this on a per-data-point basis, solving for $\sigma$ gives $\sigma = \frac{\hat{\bE} \cdot \bE}{\bE \cdot \bE}$.

\begin{figure}[ht]
    \centering
    \includegraphics[width=\columnwidth]{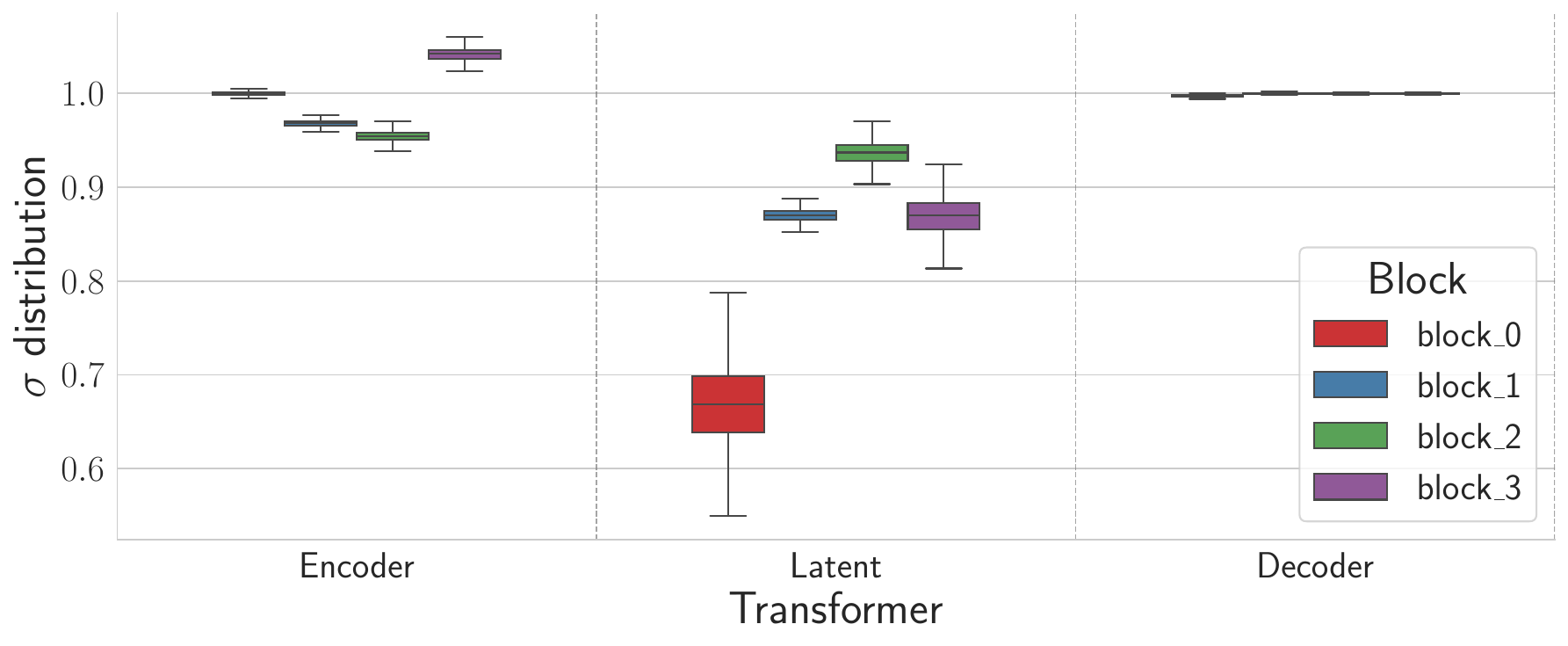}
    \includegraphics[width=\columnwidth]{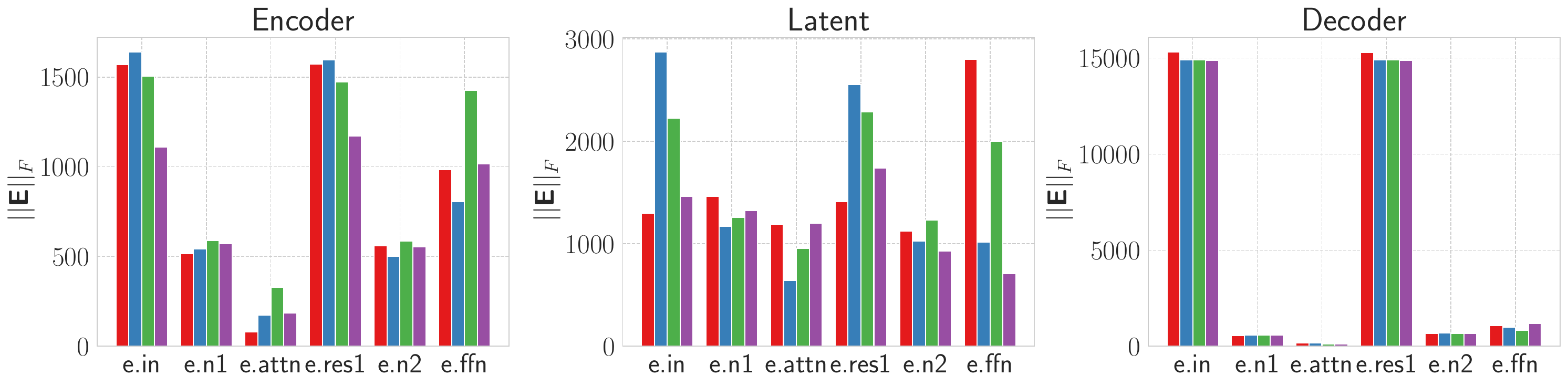}
    \caption{\textbf{Top}: $\sigma$ variation. \textbf{Bottom}: Norms of each representation inside each Transformer block. Both quantities are evaluated on the test set of \texttt{adult} dataset in the ELD-VAE architecture.}
    \label{fig:sigma}
\end{figure}

The box-plots of Fig.~\ref{fig:sigma} display the variation of $\sigma$ over the test set and across different blocks for ELD-VAE. At the decoder, the values of $\sigma$ are close to 1, indicating that the representations remain almost unchanged across the blocks. This suggests that once the model is trained, the Transformer at the decoder acts like an identity function. In the encoder, while the overall variability in representations is also low, different blocks yield distinct values of $\sigma$. On the other hand, the latent representations exhibit a higher degree of variability in $\sigma$, particularly noticeable in the first layer. To conclude this analysis, the bar-plots of Fig.~\ref{fig:sigma} show that the norm of the representation after layer normalization is lower than that of the original representation, in particular at the decoder, indicating that this layer shifts and scales the initial representation s.t. it leads to negligible representation changes.

\section{Conclusions}\label{sec:conc}

In this study, we explored the effects of applying a Transformer to the input, latent, and reconstructed representations of a VAE. Our key takeaway is a trade-off between fidelity and diversity: while using Transformers on latent and reconstructed representations increases the variability of generated data, it also reduces its fidelity. Moreover, this heightened diversity does not bring noteworthy improvements in downstream Machine-Learning tasks, raising questions about the necessity of including Transformers, given their additional computational complexity.

We further investigated the learned representations once these models were trained. A prominent observation is that the reconstructed representation processed by the Transformer essentially converges to the identity function. This behavior appears tied to residual connections, specifically because layer normalization rescales the input representation.

Future research involves examining how incorporating Transformers affects the quality of synthetic data across other generative models. A parallel priority is tailoring Transformer architectures specifically for tabular data, which presents unique challenges compared to text-based domains. Finally, given the prominent re-scaling of layer normalization, a possible direction is to study the effect of removing this component of the Transformer architecture.

\begin{ack}
This work was partially funded by projects AISym4Med (101095387) supported by Horizon Europe Cluster 1: Health, ConnectedHealth (n.o 46858), supported by Competitiveness and Internationalisation Operational Programme (POCI) and Lisbon Regional Operational Programme (LISBOA 2020), under the PORTUGAL 2020 Partnership Agreement, through the European Regional Development Fund (ERDF) and Center for Responsible AI, nr. C645008882-00000055, investment project nr. 62, financed by the Recovery and Resilience Plan (PRR) and by European Union -  NextGeneration EU. Funded by the European Union – NextGenerationEU. Views and opinions expressed are however those of the author(s) only and do not necessarily reflect those of the European Union or the European Commission. Neither the European Union nor the European Commission can be held responsible for them.
\end{ack}

\newpage


\bibliography{mybibfile}

\end{document}


\title{Supplemental Material}
\author{}
\maketitle

\tableofcontents
\newpage

\section{A Forward-Pass}\label{app:forward}

Given the introduction of embedding representations $\textbf{E}$, a detailed forward pass of the architecture that includes Transformers in all components is described. Note that in our notation, we keep the embedding dimension. From a practical perspective, this is equivalent to considering fully connected layers with weights, which are 4th order tensors. Let $\textbf{H} \in \mathbb{R}^{H\times d}$ and $\textbf{Z} \in \mathbb{R}^{L\times d}$ denote the hidden and latent representations, respectively. The following operations summarize the encoder

\begin{equation}\label{eq:ttbvae_enc}
    \begin{split}
        \textbf{H} &= \texttt{SiLU}(\fc(\ttt(\bE))) \\
        (\bmu, \blv) &= \fc(\textbf{H})
    \end{split}~~,
\end{equation}

\noindent where $\fc$ stands for a Fully-Connected Layer, and \texttt{SiLU} is the activation function~\cite{Elfwing2017SigmoidWeightedLU}. Then, we obtain $\bZ$ via the reparameterization trick, $\bZ = \bmu + \boldsymbol{\sigma} \odot \boldsymbol{\epsilon}$, where $\boldsymbol{\epsilon} \sim \mathcal{N}(\boldsymbol{0}, \textbf{I})$. In the decoding phase, a Transformer is applied to $\textbf{Z}$, and then we project it back to the hidden representation $\textbf{H}$. Finally, a representation sharing the same dimension as $\textbf{E}$ is obtained. A Transformer is applied to this representation, yielding the reconstruction embedding representation $\tilde{\textbf{E}}$. The following operations summarize the decoder

\begin{equation}\label{eq:ttbvae_dec}
    \begin{split}
        \textbf{H} &= \texttt{SiLU}(\fc(\ttt{(\textbf{Z})})) \\
        \tilde{\bE} &= \ttt(\fc(\textbf{H}))
    \end{split}~~.
\end{equation}

The reconstruction $\Tilde{\bx}$ is obtained via the Detokenization procedure described in the paper. To conclude this section, and for the sake of completeness, the model is optimized to maximize the variational lower bound

\begin{equation}\label{eq:loss}
    \mathcal{L}(\phi, \theta; \bx) = \mathbb{E}_{q_{\phi}(\bz | \bx)} \left[\log p_{\theta}(\bx | \bz)\right] - \mathbb{KL} \left[q_{\phi}(\bz | \bx) || p(\bz)\right]~~,
\end{equation}

\noindent where $\mathbb{KL}[\cdot || \cdot]$ is the Kullback-Leilbler divergence. The reconstruction loss (first term in Eq. \eqref{eq:loss}) is determined by the squared error or cross-entropy if the feature type is numerical or categorical, respectively.

\section{Metrics Evaluation}\label{app:metrics_eval}

In this section, we provide a similar analysis of Section 5.2 in the main paper for low-density estimation and machine-learning efficiency.

\begin{table}[h]
    \centering
    \caption{Average results obtained for the studied models. For a given metric, a performer with the highest average score is highlighted in \textbf{bold}, while with the lowest average rank \underline{underlined}. ($^{**}$) implies that the given model results are statistically significant with a $p$-value of 0.001 using Wilcoxon's Signed Rank Test w.r.t. the best performer.}
    \resizebox{\textwidth}{!}{%
    \begin{tabular}{l|cc|cc|cc|cc}
    & \multicolumn{4}{c|}{Low-Density} & \multicolumn{4}{c}{ML-Efficiency} \\
    \midrule
    & \multicolumn{2}{c|}{Marginals ($\uparrow$)} & \multicolumn{2}{c|}{Pairwise-Correlations ($\uparrow$)} & \multicolumn{2}{c|}{Utility ($\uparrow$)} & \multicolumn{2}{c}{ML-Fidelity ($\uparrow$)} \\
    \midrule
    & Score & Rank & Score & Rank & Score & Rank & Score & Rank \\
    Model &  &  &  &  &  &  &  &  \\
    \midrule
    VAE     & 0.913 $\pm$ \tiny{0.048} & \underline{3.20} & 0.924 $\pm$ \tiny{0.051} & 3.30 & 0.770 $\pm$ \tiny{0.174} & 3.90 & 0.805 $\pm$ \tiny{0.169} & 4.00 \\
    E-VAE   & 0.912 $\pm$ \tiny{0.047} & 3.30            & 0.921 $\pm$ \tiny{0.052} & 3.90 & 0.774 $\pm$ \tiny{0.169} & 3.60 & 0.805 $\pm$ \tiny{0.167} & 3.70 \\
    EL-VAE  & 0.910 $\pm$ \tiny{0.050} & 4.10            & 0.924 $\pm$ \tiny{0.051} & 3.60 & 0.777 $\pm$ \tiny{0.161} & 3.20 & 0.811 $\pm$ \tiny{0.158} & 3.30 \\
    ELD-VAE & 0.916 $\pm$ \tiny{0.038} & 3.60            & 0.926 $\pm$ \tiny{0.048} & 3.40 & 0.776 $\pm$ \tiny{0.174} & \underline{3.00} & 0.815 $\pm$ \tiny{0.157} & \underline{3.20} \\
    LD-VAE  & \textbf{0.917 $\pm$ \tiny{0.039}} & 3.30       & \textbf{0.928 $\pm$ \tiny{0.048}} & \underline{3.10} & \textbf{0.778 $\pm$ \tiny{0.161}} & 3.30 & \textbf{0.817 $\pm$ \tiny{0.152}} & \underline{3.20} \\
    D-VAE   & 0.903 $\pm$ \tiny{0.081} & 3.50            & 0.917 $\pm$ \tiny{0.075} & 3.70 & 0.749 $\pm$ \tiny{0.173} & 4.00 & 0.791 $\pm$ \tiny{0.160} & 3.70 \\
    \bottomrule
    \end{tabular}}%
    \label{tab:tab_main_res}
\end{table}

\subsection{Low-Density estimation}\label{subapp:low-density}

\subsubsection{Forward Sequence}\label{subsubapp:ld_fs}

Focusing on the Forward sequence (left-plots of Fig.~\ref{fig:gains_margs}), we observe little variations in all transitions for Marginal distribution estimation, except for EL-VAE $\to$ ELD-VAE, where ELD-VAE exhibits a gain of approximately 1\% for medium and large data buckets. Regarding Pairwise-Correlations, although gains variability over the considered transitions is in the range of [-0.5\%, 0.5\%], we observe a trend on the same transition, where ELD-VAE tends to produce data that is more correlated with the real one as the dataset sizes increase. Regarding the other transitions, we observe a drop in performance between the VAE and E-VAE for mid-sized datasets, while a more prominent increase in the range between E-VAE and EL-VAE.

\subsubsection{Backward Sequence}\label{subsubapp:ld_bs}

\begin{figure}
    \centering
    \includegraphics[width=0.49\textwidth]{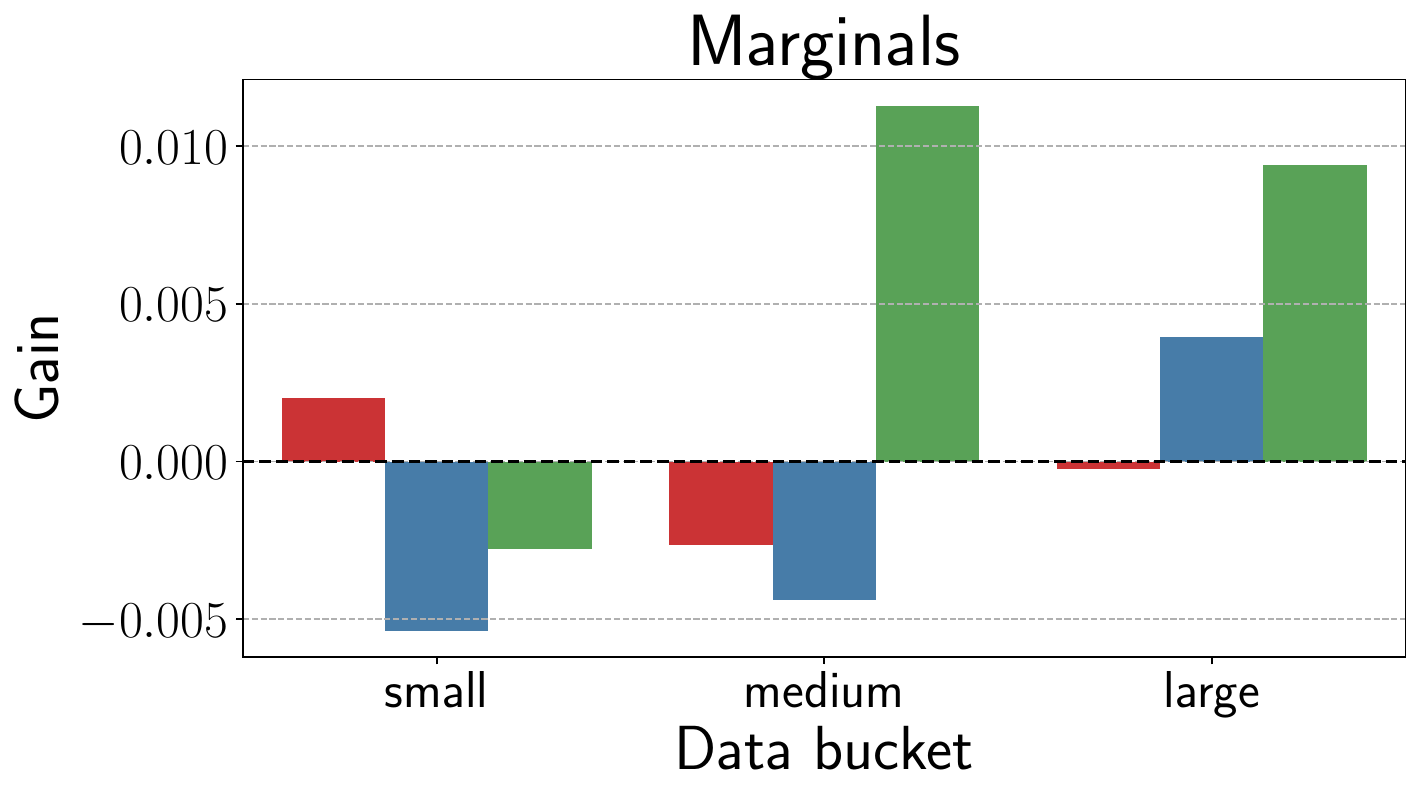}
    \includegraphics[width=0.49\textwidth]{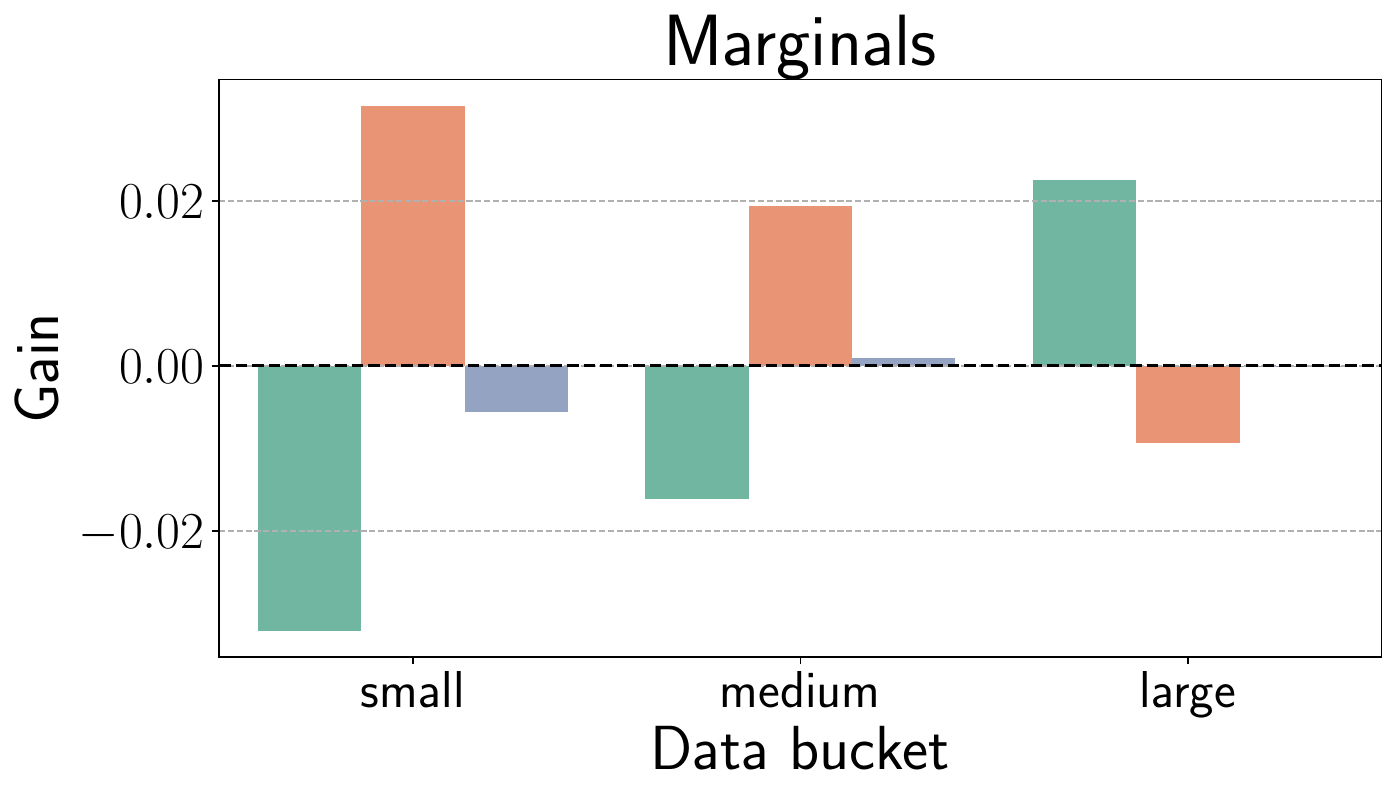}\\
    \includegraphics[width=0.49\textwidth]{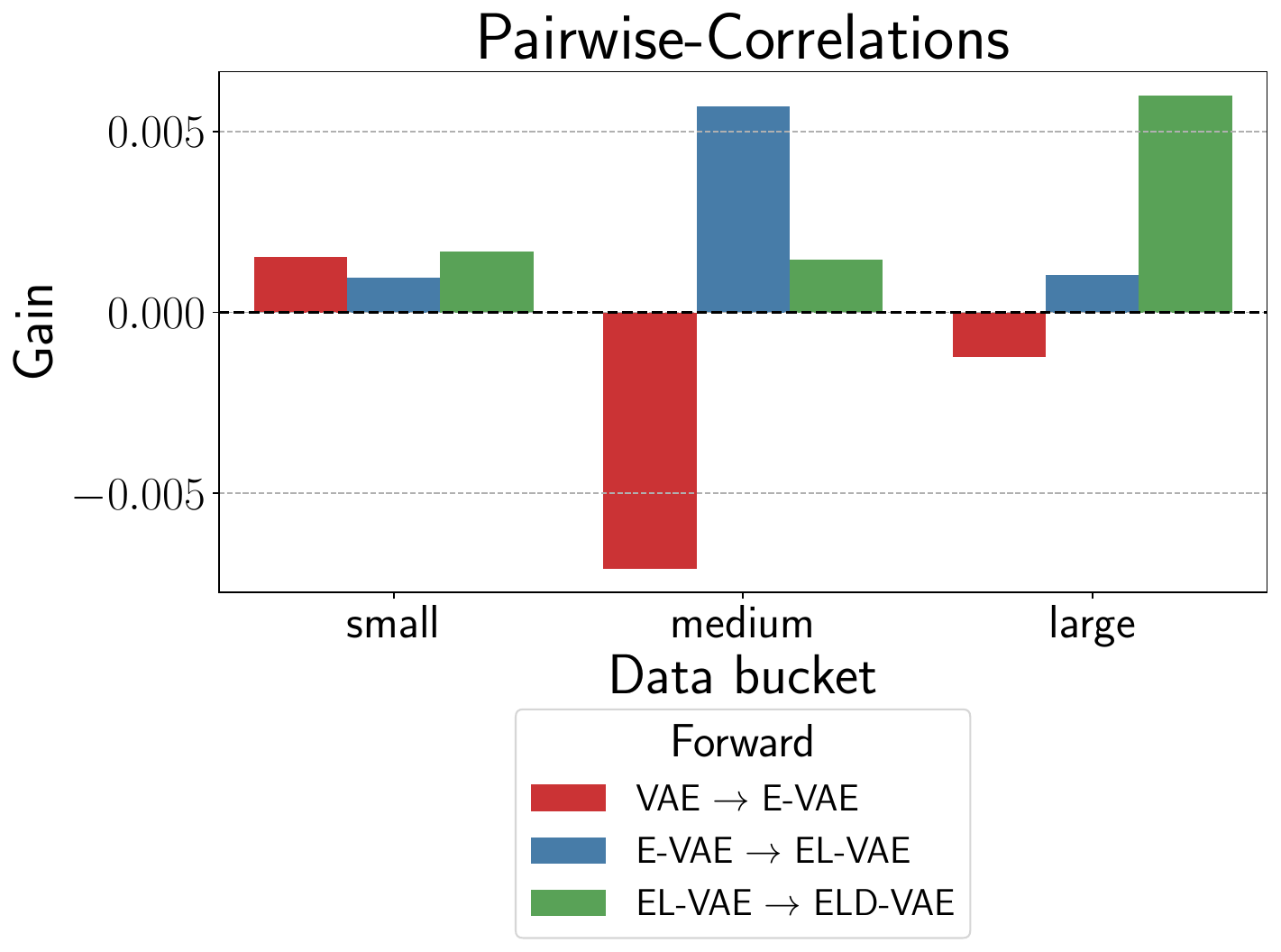}
    \includegraphics[width=0.49\textwidth]{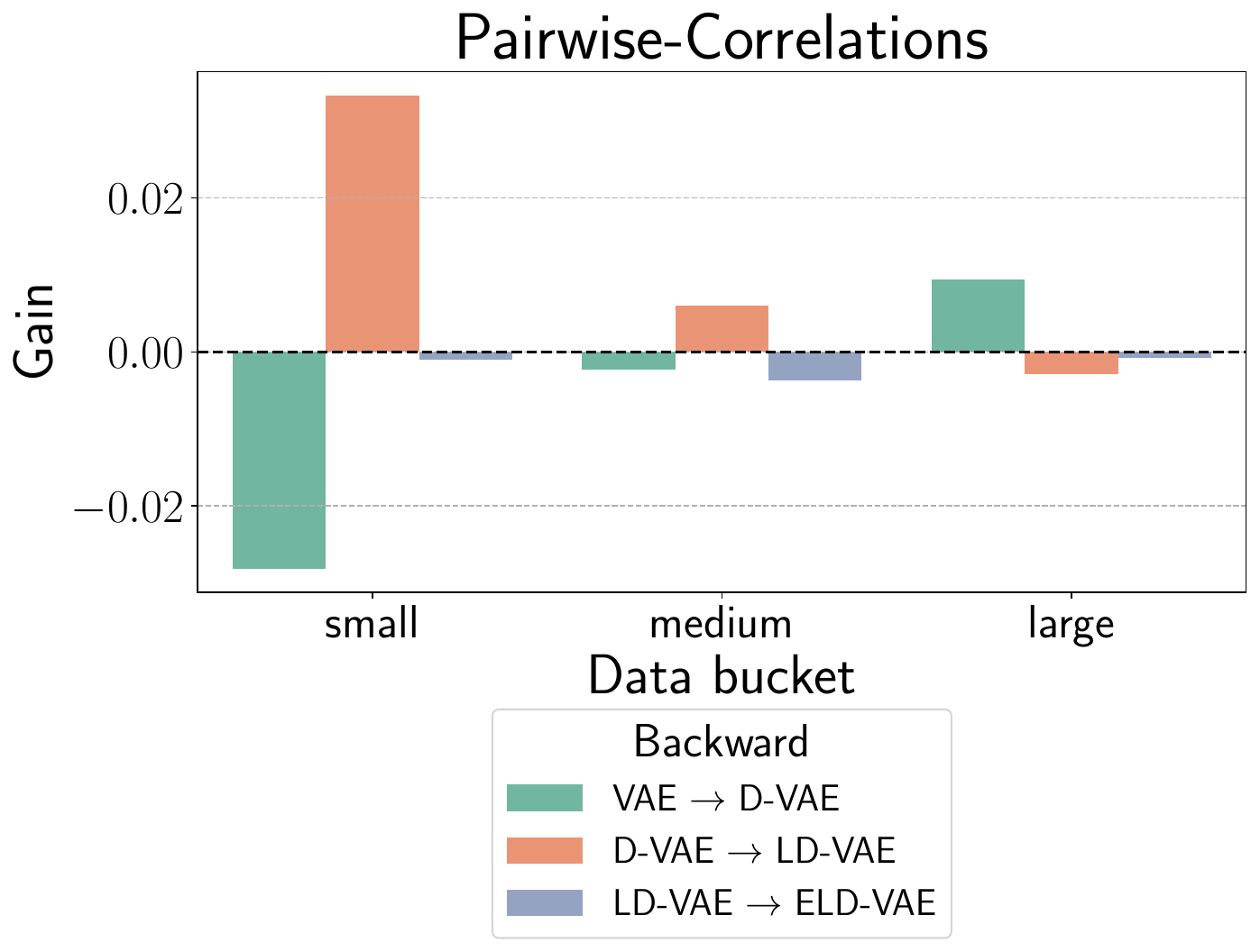}
    \caption{Aggregated gains in performance as transformers to the studied components for Marginals (top-plots) and Pairwise-Correlations (bottom-plots), for the Forward (left-plots) and Backward (right-plots) sequences.}
    \label{fig:gains_margs}
\end{figure}

Regarding the Backward sequence (right-plots of Fig.\ref{fig:gains_margs}), we observe a consistent decrease in loss of D-VAE when compared with the base implementation, as the sample size of the data increases for Marginals estimation until a gain of $\sim 2\%$ is achieved for large datasets. A reverse effect is observed between D-VAE and LD-VAE, i.e., for small and mid-size datasets LD-VAE yields better performance, while for large dataset sizes its performance deteriorates. Notably, little to no gains are observed between the transition LD-VAE $\to$ ELD-VAE, in particular for mid and large data bucket sizes. Regarding Pairwise-Correlations, the most significant losses are observed in datasets with small sizes. In particular, we observe a loss of $\sim 2\%$ in the performance of D-VAE w.r.t. VAE, while an approximate gain when considering a Transformer to model latent space representations.

\subsection{Machine Learning Efficiency}\label{app:mleff}


\subsubsection{Forward Sequence}\label{subsubapp:ml_eff_fs}

In the Forward sequence, Table~\ref{tab:tab_main_res} shows a consistent increase in average Utility scores as additional Transformers are incorporated into the network, with the exception of the transition from EL-VAE to ELD-VAE. Despite ELD-VAE exhibiting slightly lower performance relative to its predecessor, it achieves the lowest average rank within this sequence. Analyzing the top-right bar plots in Fig.~\ref{fig:gains_ml} reveals that E-VAE performance gains are most pronounced on smaller datasets, but its Utility decreases relative to the baseline for larger datasets, ultimately underperforming in that regime. By contrast, subsequent transitions exhibit the opposite trend: they incur a Utility loss for smaller datasets but yield improvements for larger ones.

Regarding ML-Fidelity, predictions obtained from models trained on synthetic data become increasingly consistent with those trained on real data as more Transformers are introduced (cf. Table~\ref{tab:tab_main_res}). This trend holds across all dataset buckets and network transitions, except for two cases: E-VAE $\to$ EL-VAE on small datasets and VAE $\to$ E-VAE on large datasets.


\subsubsection{Backward Sequence}\label{subsubapp:ml_eff_bs}

\begin{figure}
    \centering
    \includegraphics[width=0.49\textwidth]{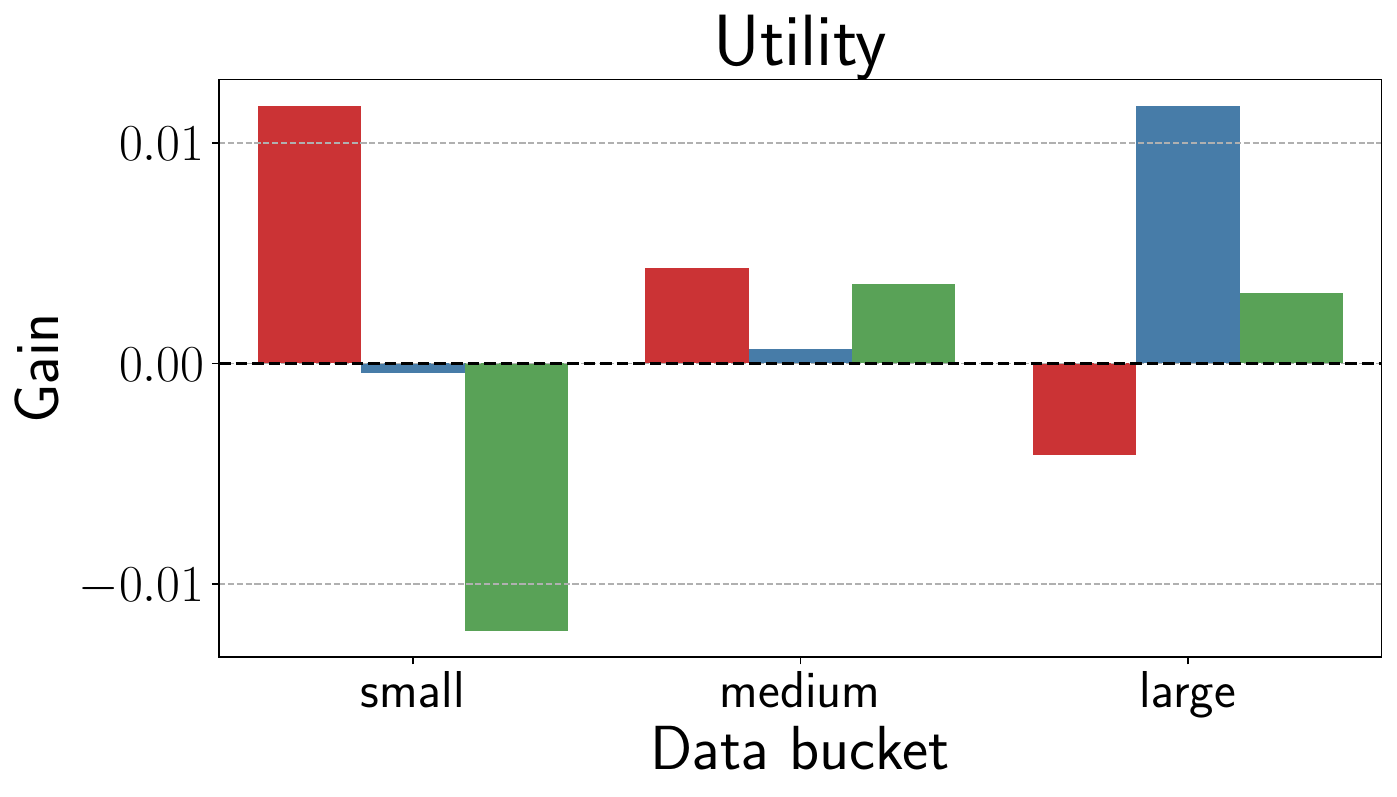}
    \includegraphics[width=0.49\textwidth]{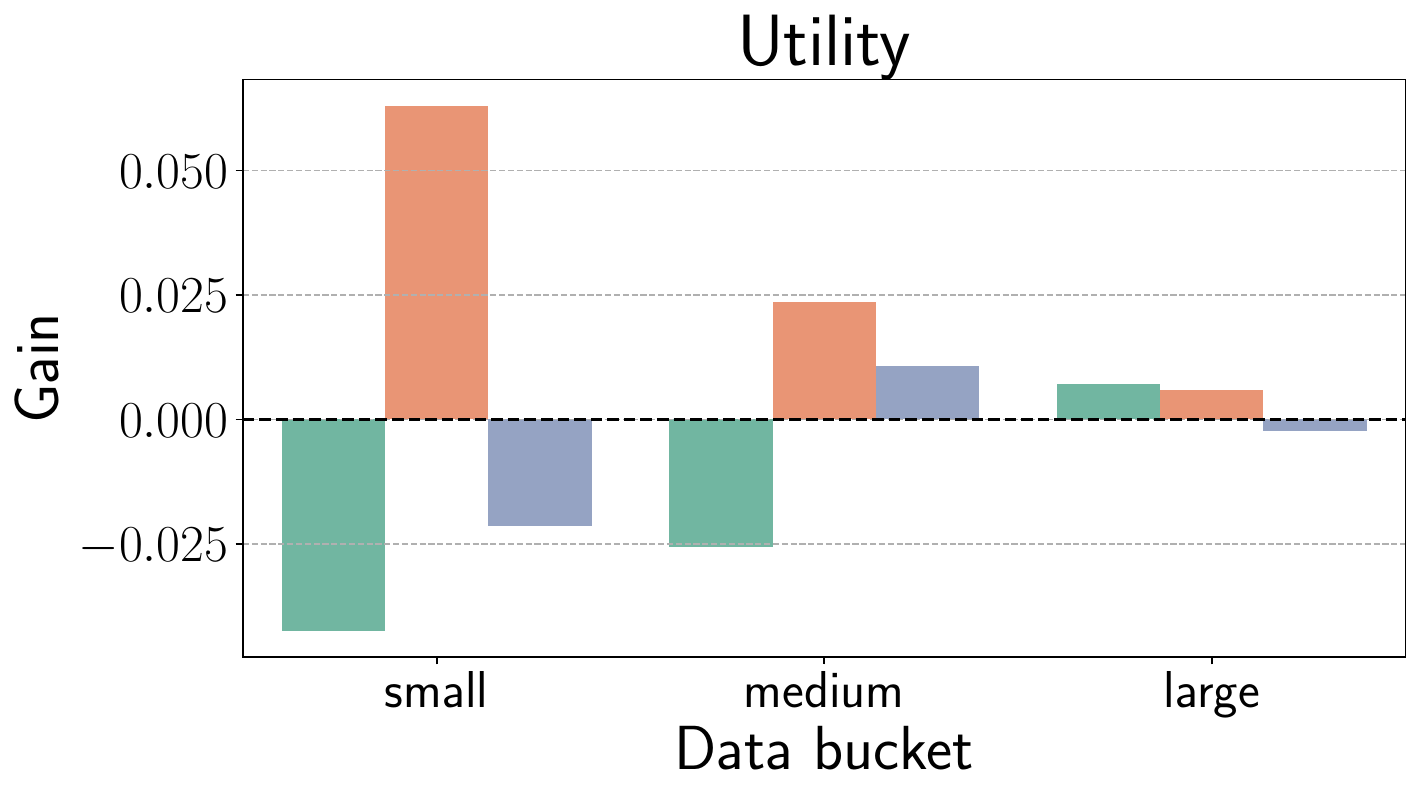}\\
    \includegraphics[width=0.49\textwidth]{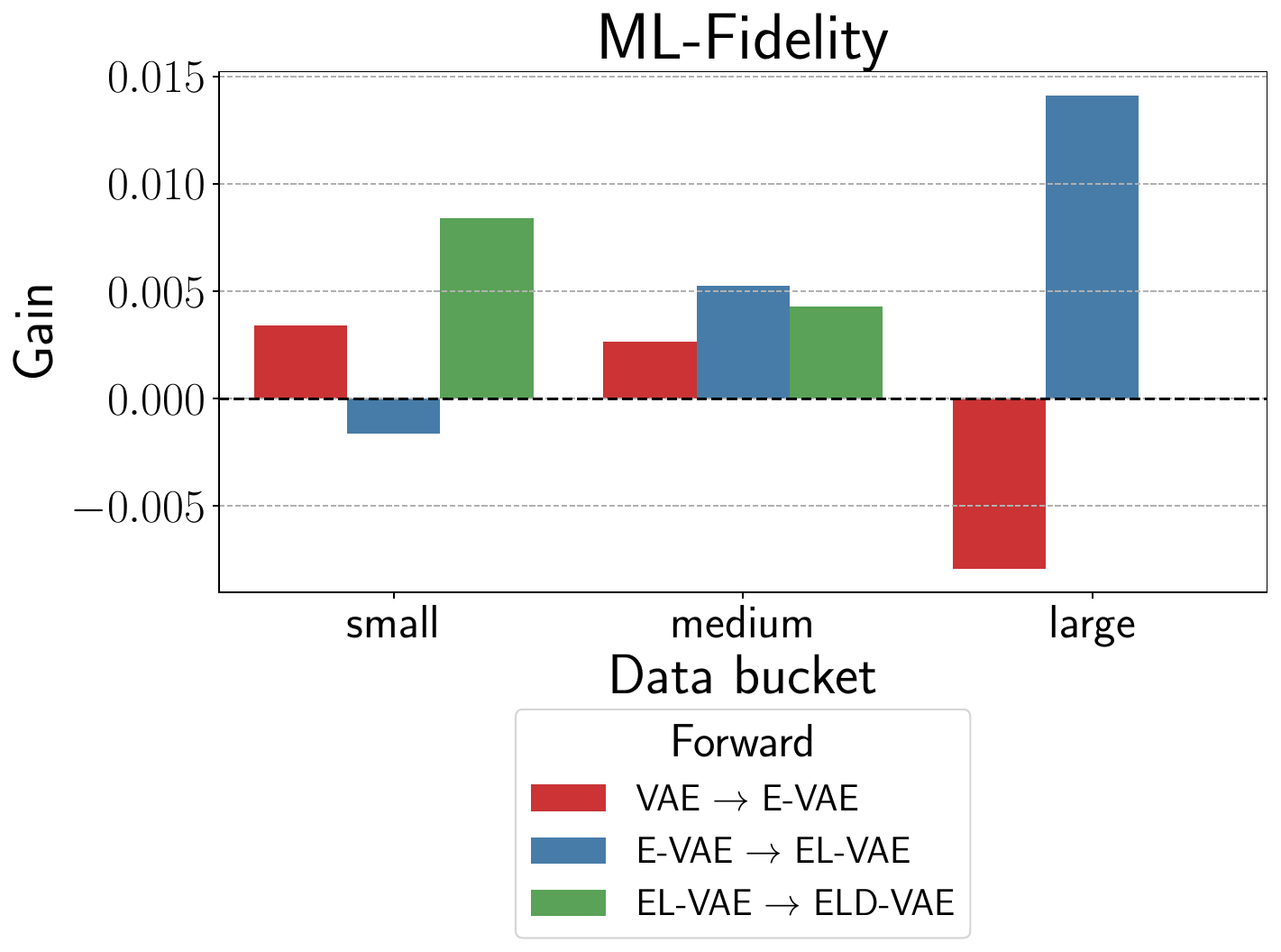}
    \includegraphics[width=0.49\textwidth]{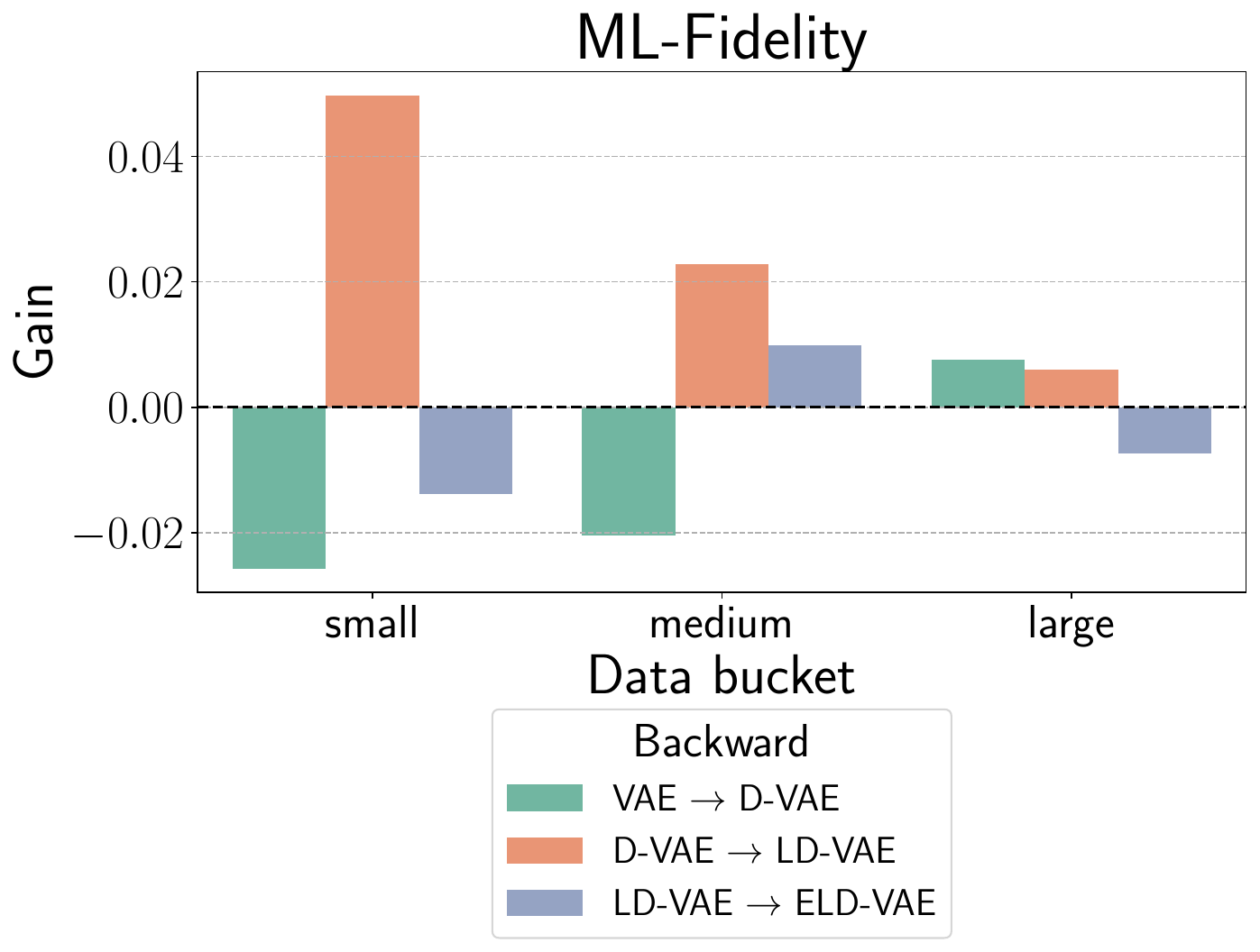}
    \caption{Aggregated gains in performance as transformers to the studied components for Utility (top-plots) and ML-Fidelity (bottom-plots), for the Forward (left-plots) and Backward (right-plots) sequences.}
    \label{fig:gains_ml}
\end{figure}

In the Backward sequence, introducing a Transformer to leverage the output representation initially reduces Utility by 2.1\%. However, the performance stabilizes quickly once a Transformer is leveraged over the latent space, where LD-VAE achieves the highest Utility. This gain is particularly evident for smaller datasets (cf. right bar plot of Fig.~\ref{fig:gains_ml}), where an improvement of 6.2\% in Utility is observed. A corresponding boost also appears in ML-Fidelity. Overall, the transitions in this sequence exhibit a consistent pattern: whenever a given transition leads to higher Utility, it also improves ML-Fidelity.

To conclude, and making a comparison between Machine-Efficiency metrics with high-density estimations, D-VAE was the model that performed poorer in terms of $\alpha$-Precision, which is related to the faithfulness of the synthetic data w.r.t. to the real one, and in terms of ML-Efficiency, although its diversity ($\beta$-Recall) was higher than the base implementation and E-VAE, which performed better in terms of ML-Efficiency. These findings suggest that, despite D-VAE ability to generate diverse representations, it does not necessarily translate into improved decision boundary estimation in terms of Utility.

\section{Sensitivity analyzes on the number of Transformer blocks}\label{app:abla}

We study the effect of lowering the number of Transformer blocks in the architecture of the Backward sequence. Three datasets are considered --- \texttt{churn}, \texttt{adult}, and \texttt{credit-approval}.

\subsection{High-density Estimation}\label{subapp:hd}

We observe that the optimal number of Transformer blocks is dataset-dependent. A clear contrast in $\alpha$-Precision for ELD-VAE is observed for \texttt{churn} and \texttt{adult} datasets, where for the first, one Transformer layer provided the best results, and then additional blocks deteriorate performance, while for the latter, the performance increases as additional blocks are considered. Notably, we observe a trade-off between $\alpha$-Precision and $\beta$-Recall as Transformer blocks are added in the ELD-VAE architecture for the \texttt{churn} dataset, where ELD-VAE yielded a higher diversity when three Transformer blocks were considered. However, this relation is not consistent with other models and datasets. 

\begin{figure}[ht]
    \centering
    \includegraphics[width=1.0\textwidth]{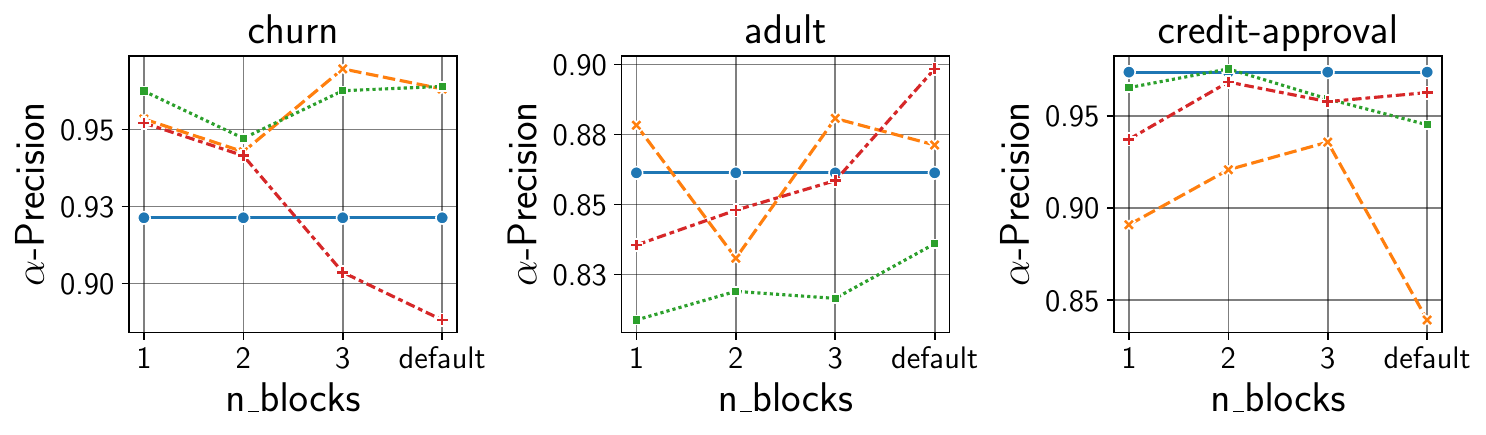}
    \includegraphics[width=1.0\textwidth]{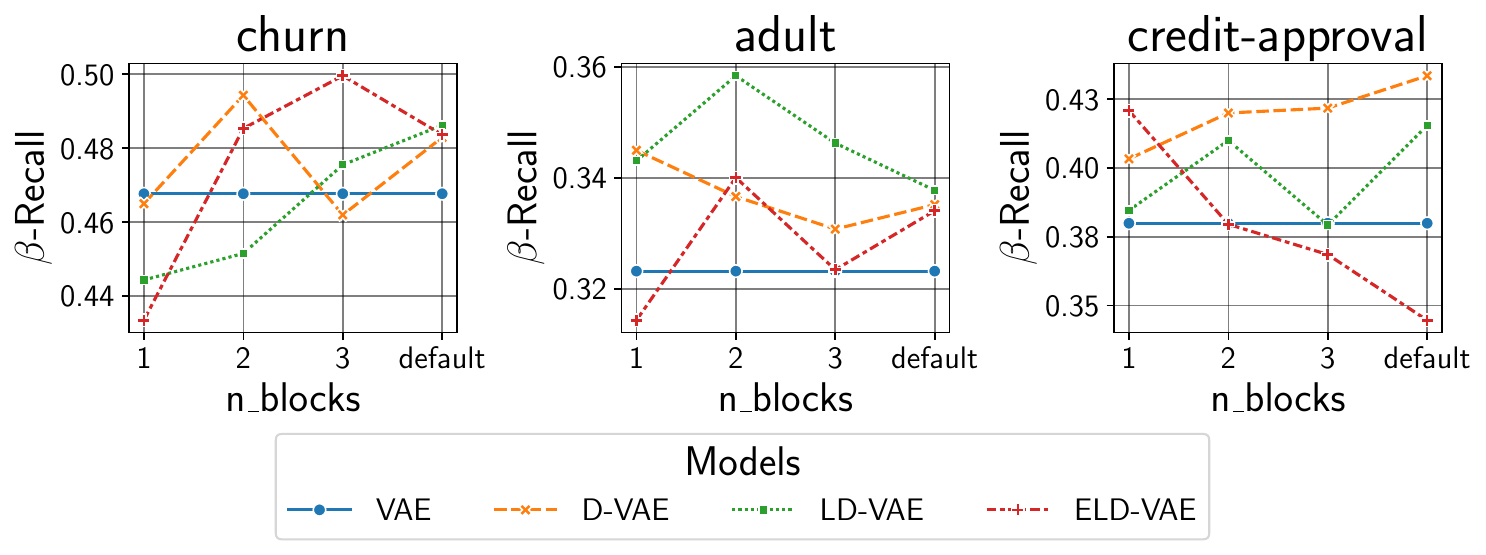}
    \caption{Effect of increasing the number of blocks in Transformers for the models of the Backward sequence.}
    \label{fig:abla}
\end{figure}

\subsection{Similarities}\label{subapp:sims}

We study the same effect of reducing the number of Transformer blocks in terms of representation similarities for ELD-VAE. We continue to observe high-similarity between representations over the same Transformer blocks that act at different components of the architecture. Notably, the most prominent effect of reducing the number of Transformer blocks seems to be present in the latent space, where we observe an increase in dissimilarity between its input representation and the remaining block representations as the number of blocks increases.

\begin{figure}[ht]
    \centering
    \includegraphics[width=\textwidth]{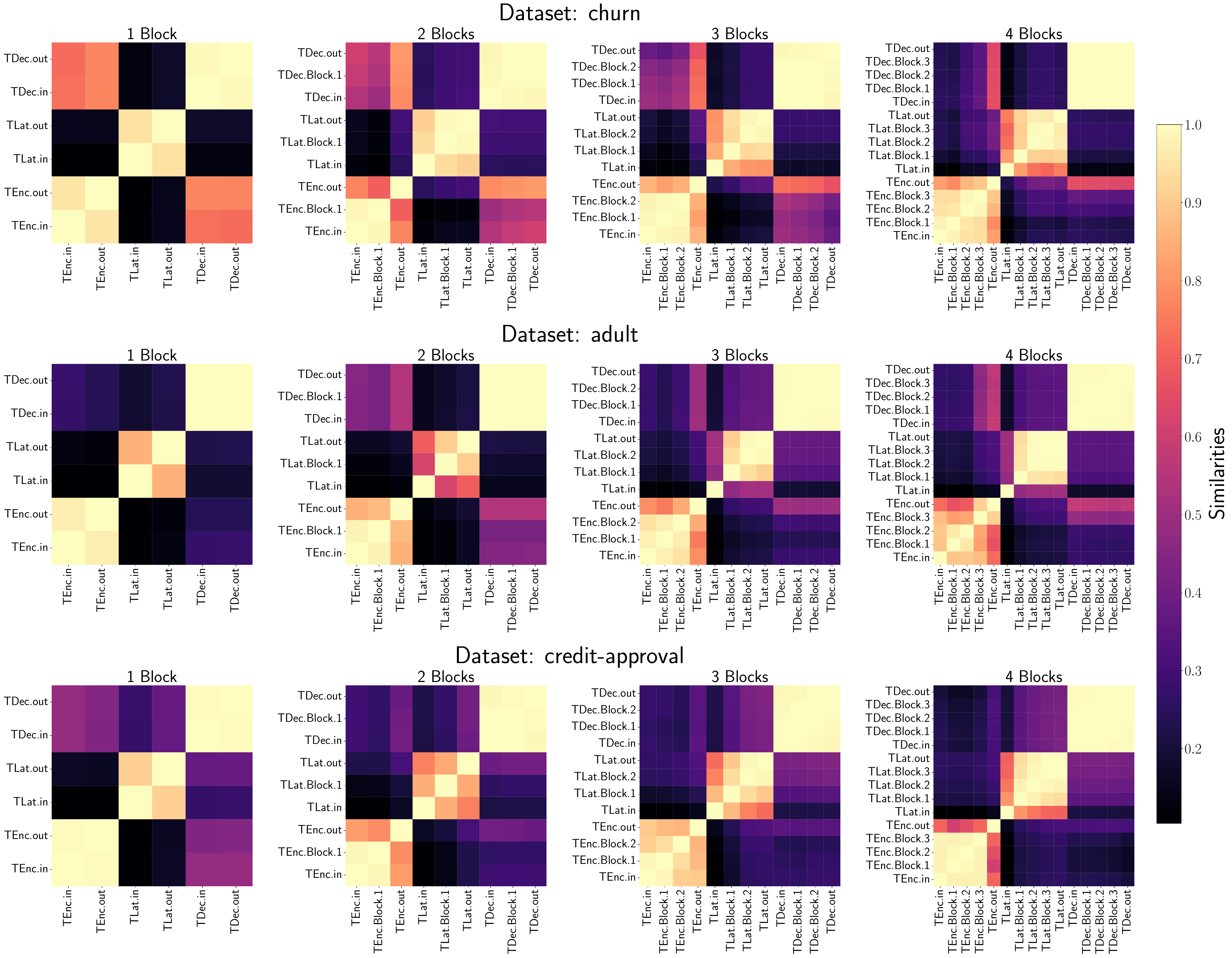}
    \caption{Similarities between representations for ELD-VAE when trained with different numbers of Transformer blocks.}
    \label{fig:sims_adult}
\end{figure}

\section{Training Instabilities}\label{app:training_instabilities}

We observed training instabilities for three datasets, across different models. These datasets are further investigated below. The loss terms depicted in Figs.~\ref{fig:kc2} and \ref{fig:jm1_and_spambase} reflect the losses obtained from the training set. 

\subsection{\texttt{kc2}}\label{subsec:kc2} 

\begin{figure}
    \centering
    \includegraphics[width=\textwidth]{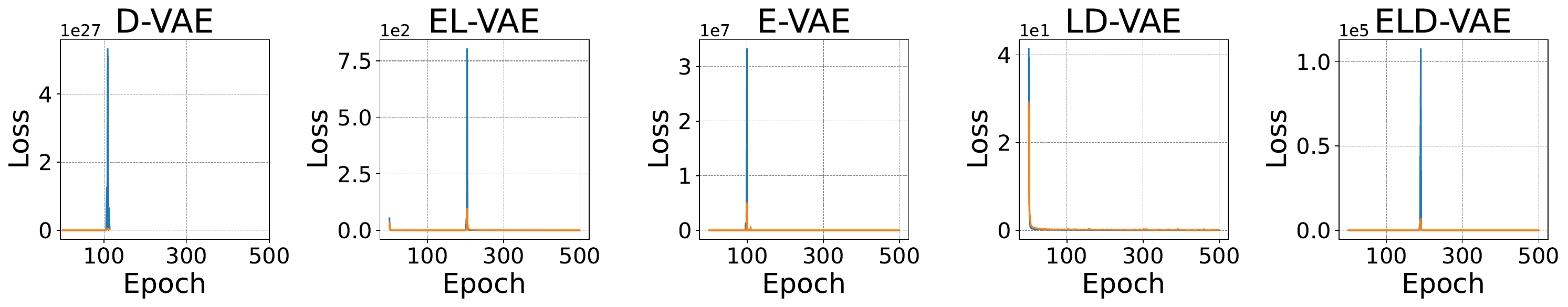}
    \includegraphics[width=\textwidth]{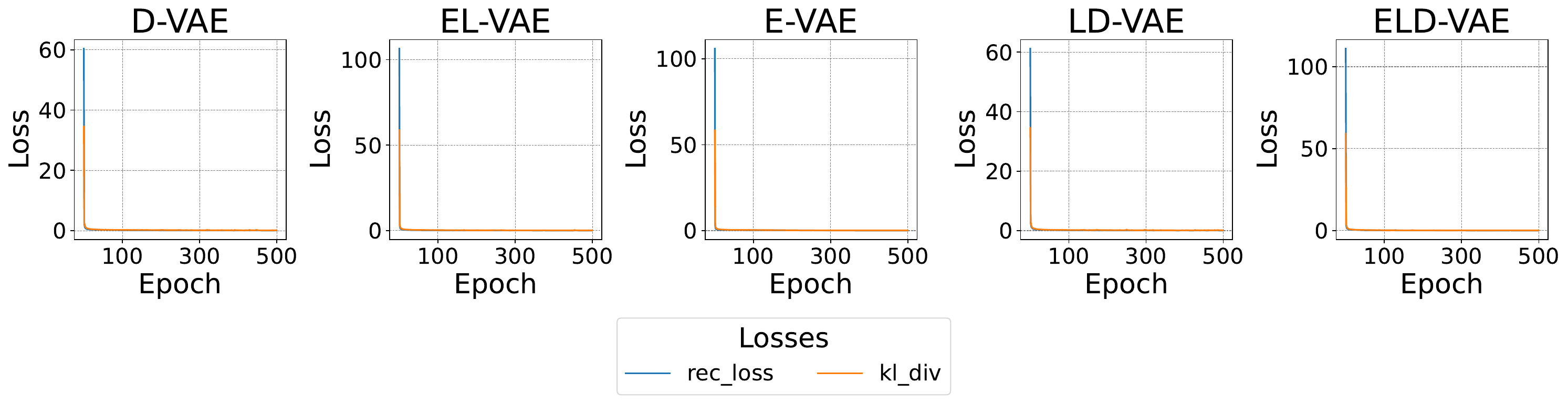}
    \caption{Losses obtained when training \texttt{kc2} dataset with default parameters (top) and half of them (bottom). We observe that the training stabilized.}
    \label{fig:kc2}
\end{figure}

It is one of the NASA Metrics Data Program defect data sets~\cite{shirabad}. It is a small dataset containing 522 samples and 22 features. Features include information about McCabe and Halstead features extractors of source code related to software quality. When training the Transformer-based models, we observed a high reconstruction loss for E-VAE, EL-VAE, LD-VAE, and ELD-VAE, while D-VAE yielded \texttt{nans} in the reconstruction loss during training (possibly due to exploding-gradient phenomena~\cite{philipp2018explodinggradientproblemdemystified}). We re-trained all models by reducing the number of hidden and latent units to half, and we observed stable training (cf.~Fig.\ref{fig:kc2}), reflecting an improvement in the High-Density estimation metrics shown in Table~\ref{tab:kc2res}, except for EL-VAE. Since we could not synthesize data from D-VAE, we excluded this dataset from our main analyses.

\begin{table}[h]
    \centering
    \caption{High-density estimations obtained for \texttt{kc2} dataset with default parameters (Before) and after reducing the hidden and latent dimensions to half.}
    \begin{tabular}{l|c|c|c|c}
    \toprule
          & \multicolumn{2}{c|}{$\alpha$-Precision} & \multicolumn{2}{c}{$\beta$-Recall}   \\
         Model & Before & After & Before & After \\
         \midrule
         D-VAE   & \texttt{nan} & 0.72 & \texttt{nan}  & 0.21 \\
         E-VAE   & 0.32 & 0.88 & 0.00  & 0.11 \\
         EL-VAE  & 0.79 & 0.62 & 0.13  & 0.17 \\
         LD-VAE  & 0.07 & 0.80 & 0.00  & 0.28 \\
         ELD-VAE & 0.54 & 0.73 & 0.07  & 0.32 \\
         \bottomrule
    \end{tabular}
    \label{tab:kc2res}
\end{table}

\subsection{\texttt{jm1}}\label{subsec:jm1}

Is another dataset from the NASA Metrics Data Program~\cite{mccabe76}. It shares almost the same features as \texttt{kc2}, except for \texttt{branchCount}, and comprises 10885 samples. We observed a peak in the reconstruction loss term at approximately half of the number of training epochs (cf. left-top Fig.~\ref{fig:jm1_and_spambase}), which was mitigated after reducing the number of parameters to half. Nevertheless, D-VAE still achieved proper results. We still re-trained the model and observed that results were poorer, as depicted in Table~\ref{tab:jm1_and_spambase}.

\subsection{\texttt{spambase}}\label{subsec:spambase}

\begin{figure}
    \centering
    \includegraphics[width=\textwidth]{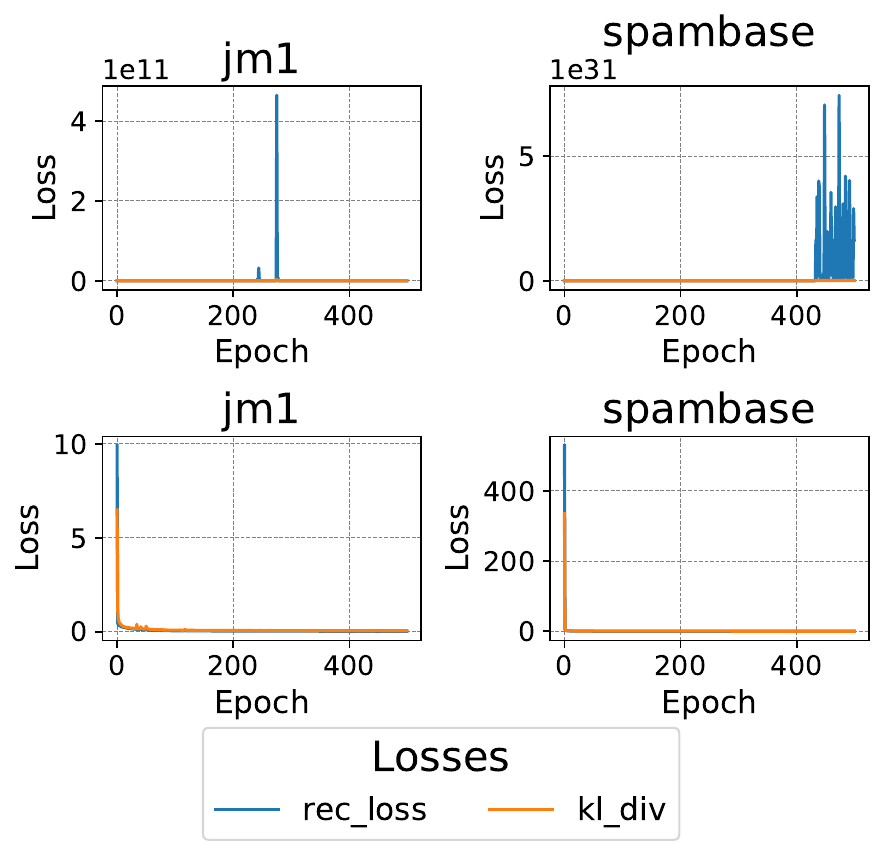}
    \caption{Training instabilities for D-VAE for \texttt{jm1} and \texttt{spambase} dataset. Training stabilized after reducing the number of parameters to half (bottom plots).}
    \label{fig:jm1_and_spambase}
\end{figure}

It is a dataset with a collection of spam e-mail statistics, such as frequencies of words. It is a dataset containing 4601 instances, with 58 continuous features. The target variable denotes if a given instance is considered spam or not based on the frequency of observed words in a given email. We observed training instabilities for D-VAE during the late stages of training (cf. top-right plot of Fig.~\ref{fig:kc2}). After training the model with half of the hidden and latent dimensions, the training went smoothly, as depicted in the bottom-right plot of Fig.~\ref{fig:jm1_and_spambase}. Initially, both $\alpha$-Precision and $\beta$-Recall were 0, while after training, these metrics yielded reasonable results (cf. Table~\ref{tab:jm1_and_spambase}).

\begin{table}[h]
    \centering
    \small
    \caption{High-density estimations obtained for D-VAE in datasets \texttt{jm1} and \texttt{spambase} with default parameters before and after reducing the hidden and latent dimensions to half.}
    \begin{tabular}{l|c|c|c|c}
    \toprule
    & \multicolumn{2}{c|}{$\alpha$-Precision} & \multicolumn{2}{c}{$\beta$-Recall} \\
    Dataset & Before & After & Before & After \\
    \midrule
    \texttt{jm1} & 0.98 & 0.90 & 0.45 & 0.28 \\
    \texttt{spambase} & 0.00 & 0.64 & 0.00 & 0.25\\
    \bottomrule
    \end{tabular}
    \label{tab:jm1_and_spambase}
\end{table}

\section{Fidelity-Diversity trade-off on the test set}\label{app:fid-div-tradeoffftt}

Given a trained model, we perform a forward-pass with the test set to produce a reconstructed version of this data. Given this reconstructed test set, we evaluate $\alpha$-Precision and $\beta$-Recall with respect to training data. Note that both datasets must share the same number of samples. Here, we perform random sampling without replacement over the training data. The averaged results are presented in Table~\ref{tab:res_test}.

\begin{table}[h]
    \centering
    \small
    \caption{Averaged results for high-density estimation metrics over the test-set for the considered models. A model with the highest average score in \textbf{bold}, while the model with the lowest average rank in \underline{underlined}.}
    \begin{tabular}{l|cc|cc}
        \toprule
         & \multicolumn{2}{c|}{$\alpha$-Precision} & \multicolumn{2}{c}{$\beta$-Recall} \\
         & Score & Rank & Score & Rank \\
        Model &  &  &  &  \\
        \midrule
        VAE & 0.902 $\pm$ \tiny{0.105} & 3.50 & 0.483 $\pm$ \tiny{0.162} & 5.00 \\
        E-VAE & \textbf{0.905 $\pm$ \tiny{0.100}} & 3.60 & 0.499 $\pm$ \tiny{0.169} & 4.30 \\
        EL-VAE & 0.905 $\pm$ \tiny{0.112} & \underline{3.10} & 0.543 $\pm$ \tiny{0.160} & 2.80 \\
        ELD-VAE & 0.894 $\pm$ \tiny{0.137} & 3.60 & \textbf{0.555 $\pm$ \tiny{0.153}} & \underline{2.30} \\
        LD-VAE & 0.892 $\pm$ \tiny{0.141} & 3.80 & 0.549 $\pm$ \tiny{0.153} & 2.80 \\
        D-VAE & 0.870 $\pm$ \tiny{0.201} & 3.40 & 0.510 $\pm$ \tiny{0.185} & 3.80 \\
        \bottomrule
    \end{tabular}    
    \label{tab:res_test}
\end{table}

Notably, in terms of $\alpha$-Precision, all models compete with each other, except for LD-VAE and D-VAE, which had the lowest score. Nevertheless, D-VAE ascertained a higher rank than ELD-VAE. On the other hand, in terms of $\beta$-Recall, similar conclusions concerning previous ones in the main paper are drawn, i.e., diversity increases with the introduction of Transformers to model representations at the latent and decoder levels.

\newpage

\section{Datasets}\label{app:datasets}

\begin{table}[ht]
    \centering
    \caption{Datasets with their main properties. \# Train, Val., and Test represent the number of instances in the Training, Validation, and Test set, while \# Num. and Cat. represent the number of numerical and categorical features, respectively.}
    \footnotesize
    \resizebox{0.70\textwidth}{!}{
    \begin{tabular}{rlrrrrrl}
\toprule
Id & dataname & \# Train & \# Val. & \# Test & \# Num. & \# Cat. & Task \\
\midrule
3 & kr-vs-kp & 2444 & 320 & 432 & 0 & 36 & binary \\
6 & letter & 15300 & 2000 & 2700 & 16 & 0 & multi-class \\
11 & balance-scale & 477 & 63 & 85 & 4 & 0 & multi-class \\
12 & mfeat-factors & 1530 & 200 & 270 & 216 & 0 & multi-class \\
14 & mfeat-fourier & 1530 & 200 & 270 & 76 & 0 & multi-class \\
15 & breast-w & 534 & 70 & 95 & 9 & 0 & binary \\
16 & mfeat-karhunen & 1530 & 200 & 270 & 64 & 0 & multi-class \\
18 & mfeat-morphological & 1530 & 200 & 270 & 6 & 0 & multi-class \\
22 & mfeat-zernike & 1530 & 200 & 270 & 47 & 0 & multi-class \\
23 & cmc & 1126 & 148 & 199 & 2 & 7 & multi-class \\
29 & credit-approval & 527 & 69 & 94 & 6 & 9 & binary \\
31 & credit-g & 765 & 100 & 135 & 7 & 13 & binary \\
37 & diabetes & 587 & 77 & 104 & 8 & 0 & binary \\
38 & sick & 2884 & 378 & 510 & 7 & 22 & binary \\
44 & spambase & 3519 & 461 & 621 & 57 & 0 & binary \\
46 & splice & 2440 & 319 & 431 & 0 & 60 & multi-class \\
50 & tic-tac-toe & 732 & 96 & 130 & 0 & 9 & binary \\
54 & vehicle & 646 & 85 & 115 & 18 & 0 & multi-class \\
151 & electricity & 34663 & 4532 & 6117 & 7 & 1 & binary \\
182 & satimage & 4918 & 643 & 869 & 36 & 0 & multi-class \\
188 & eucalyptus & 562 & 74 & 100 & 14 & 5 & multi-class \\
307 & vowel & 757 & 99 & 134 & 10 & 2 & multi-class \\
458 & analcatdata\_authorship & 642 & 85 & 114 & 70 & 0 & multi-class \\
469 & analcatdata\_dmft & 609 & 80 & 108 & 0 & 4 & multi-class \\
1049 & pc4 & 1115 & 146 & 197 & 37 & 0 & binary \\
1050 & pc3 & 1195 & 157 & 211 & 37 & 0 & binary \\
1053 & jm1 & 8326 & 1089 & 1470 & 21 & 0 & binary \\
1067 & kc1 & 1613 & 211 & 285 & 21 & 0 & binary \\
1068 & pc1 & 848 & 111 & 150 & 21 & 0 & binary \\
1461 & bank-marketing & 34585 & 4522 & 6104 & 7 & 9 & binary \\
1462 & banknote-authentication & 1048 & 138 & 186 & 4 & 0 & binary \\
1464 & blood-transfusion-service-center & 572 & 75 & 101 & 4 & 0 & binary \\
1475 & first-order-theorem-proving & 4680 & 612 & 826 & 51 & 0 & multi-class \\
1480 & ilpd & 445 & 59 & 79 & 9 & 1 & binary \\
1486 & nomao & 26365 & 3447 & 4653 & 89 & 29 & binary \\
1487 & ozone-level-8hr & 1938 & 254 & 342 & 72 & 0 & binary \\
1489 & phoneme & 4133 & 541 & 730 & 5 & 0 & binary \\
1497 & wall-robot-navigation & 4173 & 546 & 737 & 24 & 0 & multi-class \\
1510 & wdbc & 435 & 57 & 77 & 30 & 0 & binary \\
1590 & adult & 37363 & 4885 & 6594 & 6 & 8 & binary \\
4534 & PhishingWebsites & 8456 & 1106 & 1493 & 0 & 30 & binary \\
4538 & GesturePhaseSegmentationProcessed & 7552 & 988 & 1333 & 32 & 0 & multi-class \\
6332 & cylinder-bands & 413 & 54 & 73 & 18 & 19 & binary \\
23381 & dresses-sales & 382 & 50 & 68 & 1 & 11 & binary \\
40499 & texture & 4207 & 550 & 743 & 40 & 0 & multi-class \\
40668 & connect-4 & 51680 & 6756 & 9121 & 0 & 42 & multi-class \\
40670 & dna & 2436 & 319 & 431 & 0 & 180 & multi-class \\
40701 & churn & 3825 & 500 & 675 & 16 & 4 & binary \\
40966 & MiceProtein & 826 & 108 & 146 & 77 & 0 & multi-class \\
40975 & car & 1321 & 173 & 234 & 0 & 6 & multi-class \\
40979 & mfeat-pixel & 1530 & 200 & 270 & 240 & 0 & multi-class \\
40982 & steel-plates-fault & 1484 & 195 & 262 & 27 & 0 & multi-class \\
40983 & wilt & 3701 & 484 & 654 & 5 & 0 & binary \\
40984 & segment & 1767 & 231 & 312 & 16 & 0 & multi-class \\
40994 & climate-model-simulation-crashes & 413 & 54 & 73 & 18 & 0 & binary \\
41027 & jungle\_chess\_2pcs\_raw\_endgame\_complete & 34286 & 4482 & 6051 & 6 & 0 & multi-class \\
\bottomrule
\end{tabular}
}
    \label{tab:datasets}
\end{table}

\newpage

\section{Results}\label{app:results}

\subsection{Low-Density}\label{subapp:low_density}

\begin{table*}[ht]
    \centering
    \caption{Low-Density scores obtained for each model. Results in \textbf{bold} highlights the top-performer.}
    \scriptsize
    \resizebox{0.8\textwidth}{!}{
    \begin{tabular}{l|cccccc|ccccccc}
\toprule
 & \multicolumn{6}{c|}{Marginals} & \multicolumn{6}{c}{Pairwise-Correlations} \\
Id & & & & & & & & & & & & \\
 & VAE & E-VAE & EL-VAE & ELD-VAE & LD-VAE & D-VAE & VAE & E-VAE & EL-VAE & ELD-VAE & LD-VAE & D-VAE\\
\midrule
3 & 0.978 & 0.976 & \textbf{0.983} & 0.979 & 0.979 & 0.978 & 0.963 & 0.959 & \textbf{0.970} & 0.964 & 0.962 & 0.960 \\
6 & \textbf{0.954} & 0.946 & 0.947 & 0.947 & 0.947 & 0.948 & \textbf{0.956} & 0.953 & 0.952 & 0.951 & 0.953 & 0.950 \\
11 & 0.883 & \textbf{0.884} & 0.850 & 0.875 & 0.863 & 0.854 & 0.865 & 0.877 & 0.853 & 0.888 & 0.883 & \textbf{0.889} \\
12 & 0.900 & 0.904 & 0.908 & \textbf{0.916} & 0.907 & 0.910 & 0.972 & 0.969 & \textbf{0.974} & 0.974 & 0.974 & 0.972 \\
14 & 0.891 & 0.891 & 0.879 & 0.884 & 0.891 & \textbf{0.892} & 0.970 & \textbf{0.973} & 0.964 & 0.963 & 0.967 & 0.970 \\
15 & 0.805 & 0.816 & 0.757 & \textbf{0.836} & 0.822 & 0.605 & 0.759 & 0.746 & 0.756 & \textbf{0.857} & 0.831 & 0.636 \\
16 & 0.912 & \textbf{0.916} & 0.906 & 0.906 & 0.906 & 0.912 & \textbf{0.976} & 0.976 & 0.973 & 0.973 & 0.973 & 0.973 \\
18 & 0.864 & 0.870 & 0.856 & 0.884 & \textbf{0.900} & 0.885 & 0.793 & 0.786 & 0.816 & 0.823 & \textbf{0.838} & 0.785 \\
22 & 0.943 & \textbf{0.946} & 0.931 & 0.934 & 0.929 & 0.932 & 0.972 & \textbf{0.973} & 0.973 & 0.973 & 0.971 & 0.969 \\
23 & \textbf{0.969} & 0.957 & 0.963 & 0.969 & 0.958 & 0.946 & 0.904 & 0.893 & 0.902 & \textbf{0.907} & 0.902 & 0.887 \\
29 & 0.917 & 0.906 & 0.901 & 0.907 & \textbf{0.918} & 0.913 & \textbf{0.917} & 0.908 & 0.898 & 0.881 & 0.913 & 0.899 \\
31 & 0.933 & 0.927 & 0.912 & 0.935 & \textbf{0.945} & 0.930 & \textbf{0.874} & 0.860 & 0.848 & 0.858 & 0.869 & 0.855 \\
37 & 0.873 & 0.880 & 0.870 & 0.849 & 0.886 & \textbf{0.896} & 0.934 & 0.928 & 0.940 & 0.933 & 0.931 & \textbf{0.948} \\
38 & 0.959 & \textbf{0.966} & 0.961 & 0.959 & 0.964 & 0.964 & 0.933 & 0.942 & 0.926 & 0.930 & \textbf{0.962} & 0.928 \\
44 & 0.870 & 0.879 & 0.899 & \textbf{0.931} & 0.931 & 0.512 & 0.969 & 0.968 & 0.968 & 0.973 & \textbf{0.977} & 0.916 \\
46 & \textbf{0.964} & 0.958 & 0.922 & 0.893 & 0.919 & 0.950 & \textbf{0.939} & 0.931 & 0.885 & 0.854 & 0.884 & 0.922 \\
50 & 0.972 & 0.970 & 0.969 & 0.965 & \textbf{0.975} & 0.964 & 0.941 & 0.943 & 0.937 & 0.933 & \textbf{0.945} & 0.937 \\
54 & 0.907 & 0.899 & \textbf{0.915} & 0.895 & 0.900 & 0.881 & 0.949 & 0.937 & 0.948 & 0.950 & \textbf{0.952} & 0.935 \\
151 & 0.937 & 0.933 & 0.932 & 0.955 & 0.962 & \textbf{0.963} & 0.941 & 0.937 & 0.939 & 0.955 & 0.951 & \textbf{0.958} \\
182 & 0.943 & \textbf{0.948} & 0.940 & 0.942 & 0.941 & 0.930 & 0.929 & 0.933 & 0.936 & \textbf{0.946} & 0.945 & 0.935 \\
188 & 0.879 & \textbf{0.883} & 0.863 & 0.870 & 0.861 & 0.837 & 0.794 & 0.789 & \textbf{0.801} & 0.787 & 0.773 & 0.747 \\
307 & 0.919 & \textbf{0.922} & 0.913 & 0.898 & 0.910 & 0.922 & 0.895 & \textbf{0.896} & 0.890 & 0.877 & 0.881 & 0.890 \\
458 & 0.839 & \textbf{0.843} & 0.834 & 0.828 & 0.835 & 0.832 & 0.955 & 0.961 & 0.964 & \textbf{0.965} & 0.963 & 0.961 \\
469 & 0.942 & 0.936 & \textbf{0.959} & 0.948 & 0.932 & 0.789 & 0.870 & 0.875 & \textbf{0.889} & 0.877 & 0.864 & 0.687 \\
1049 & 0.819 & 0.797 & 0.824 & \textbf{0.86} & 0.848 & 0.858 & 0.939 & 0.923 & 0.943 & 0.931 & 0.944 & \textbf{0.946} \\
1050 & 0.836 & 0.820 & 0.820 & 0.863 & \textbf{0.893} & 0.865 & \textbf{0.935} & 0.923 & 0.929 & 0.922 & 0.923 & 0.922 \\
1053 & 0.849 & 0.849 & 0.905 & 0.906 & 0.890 & \textbf{0.957} & 0.916 & 0.897 & 0.897 & 0.924 & 0.904 & \textbf{0.943} \\
1067 & 0.818 & 0.833 & 0.786 & \textbf{0.859} & 0.849 & 0.844 & 0.847 & 0.856 & 0.859 & 0.925 & 0.939 & \textbf{0.944} \\
1068 & 0.850 & 0.837 & 0.858 & \textbf{0.891} & 0.859 & 0.839 & 0.912 & 0.916 & 0.932 & \textbf{0.952} & 0.942 & 0.945 \\
1461 & 0.939 & 0.938 & 0.935 & 0.947 & \textbf{0.958} & 0.954 & 0.927 & 0.924 & 0.925 & 0.923 & \textbf{0.932} & 0.930 \\
1462 & \textbf{0.947} & 0.936 & 0.929 & 0.931 & 0.927 & 0.935 & 0.931 & 0.921 & 0.923 & 0.925 & 0.919 & \textbf{0.932} \\
1464 & 0.855 & 0.877 & \textbf{0.916} & 0.870 & 0.909 & 0.913 & 0.899 & 0.908 & 0.914 & 0.917 & 0.924 & \textbf{0.927} \\
1475 & 0.854 & 0.861 & 0.862 & 0.890 & 0.880 & \textbf{0.897} & 0.940 & 0.946 & 0.951 & 0.950 & 0.953 & \textbf{0.956} \\
1480 & \textbf{0.927} & 0.924 & 0.921 & 0.892 & 0.924 & 0.901 & 0.942 & \textbf{0.945} & 0.945 & 0.928 & 0.944 & 0.945 \\
1486 & 0.905 & 0.898 & 0.934 & 0.943 & 0.938 & \textbf{0.944} & 0.897 & 0.898 & 0.916 & \textbf{0.92} & 0.919 & 0.919 \\
1487 & \textbf{0.941} & 0.937 & 0.935 & 0.931 & 0.935 & 0.937 & \textbf{0.984} & 0.983 & 0.982 & 0.982 & 0.980 & 0.984 \\
1489 & 0.961 & \textbf{0.962} & 0.961 & 0.947 & 0.961 & 0.960 & 0.939 & 0.941 & 0.935 & 0.937 & \textbf{0.945} & 0.935 \\
1497 & 0.910 & 0.908 & 0.908 & 0.901 & 0.917 & \textbf{0.921} & 0.959 & 0.954 & 0.956 & 0.958 & \textbf{0.963} & 0.958 \\
1510 & 0.895 & 0.903 & 0.905 & \textbf{0.927} & 0.921 & 0.902 & 0.948 & 0.963 & \textbf{0.969} & 0.965 & 0.965 & 0.960 \\
1590 & 0.944 & 0.946 & 0.942 & 0.942 & 0.944 & \textbf{0.947} & 0.901 & 0.901 & \textbf{0.902} & 0.902 & 0.897 & 0.900 \\
4534 & \textbf{0.984} & 0.984 & 0.974 & 0.971 & 0.966 & 0.978 & 0.971 & \textbf{0.972} & 0.958 & 0.953 & 0.946 & 0.962 \\
4538 & \textbf{0.958} & 0.958 & 0.943 & 0.951 & 0.944 & 0.951 & \textbf{0.973} & 0.971 & 0.967 & 0.967 & 0.970 & 0.972 \\
6332 & 0.900 & \textbf{0.902} & 0.895 & 0.880 & 0.888 & 0.884 & 0.793 & \textbf{0.808} & 0.793 & 0.767 & 0.789 & 0.743 \\
23381 & \textbf{0.938} & 0.934 & 0.926 & 0.909 & 0.893 & 0.841 & \textbf{0.837} & 0.835 & 0.835 & 0.831 & 0.817 & 0.737 \\
40499 & 0.933 & 0.937 & 0.918 & 0.938 & 0.930 & \textbf{0.944} & 0.973 & 0.974 & 0.969 & 0.978 & 0.976 & \textbf{0.979} \\
40668 & \textbf{0.986} & 0.985 & 0.986 & 0.982 & 0.985 & 0.985 & 0.974 & 0.973 & \textbf{0.975} & 0.968 & 0.973 & 0.972 \\
40670 & 0.965 & 0.966 & \textbf{0.971} & 0.963 & 0.964 & 0.957 & 0.945 & 0.947 & \textbf{0.954} & 0.941 & 0.942 & 0.932 \\
40701 & 0.950 & 0.951 & 0.949 & 0.952 & \textbf{0.964} & 0.960 & 0.953 & 0.957 & 0.950 & 0.956 & \textbf{0.968} & 0.960 \\
40966 & 0.902 & 0.902 & 0.889 & 0.905 & 0.890 & \textbf{0.908} & 0.964 & 0.965 & 0.962 & 0.965 & \textbf{0.966} & 0.960 \\
40975 & \textbf{0.963} & 0.957 & 0.957 & 0.941 & 0.934 & 0.923 & \textbf{0.93} & 0.923 & 0.925 & 0.910 & 0.902 & 0.882 \\
40979 & 0.864 & 0.866 & 0.855 & \textbf{0.879} & 0.850 & 0.849 & 0.945 & 0.943 & 0.942 & \textbf{0.946} & 0.944 & 0.944 \\
40982 & \textbf{0.907} & 0.894 & 0.885 & 0.891 & 0.897 & 0.889 & \textbf{0.944} & 0.939 & 0.942 & 0.938 & 0.939 & 0.941 \\
40983 & 0.970 & 0.955 & \textbf{0.971} & 0.968 & 0.966 & 0.955 & 0.954 & 0.863 & \textbf{0.967} & 0.958 & 0.960 & 0.953 \\
40984 & 0.919 & \textbf{0.932} & 0.915 & 0.916 & 0.921 & 0.928 & 0.930 & 0.923 & 0.932 & 0.939 & 0.936 & \textbf{0.95} \\
40994 & 0.945 & \textbf{0.95} & 0.948 & 0.947 & 0.947 & 0.948 & 0.965 & 0.966 & 0.970 & \textbf{0.971} & 0.970 & 0.970 \\
41027 & 0.816 & 0.815 & 0.844 & 0.916 & 0.904 & \textbf{0.943} & 0.884 & 0.882 & 0.900 & 0.937 & 0.941 & \textbf{0.952} \\
\bottomrule
\end{tabular}
}
    \label{tab:lowdensity}
\end{table*}

\newpage

\subsection{High-Density}\label{subapp:hd_res}

\begin{table*}[h]
    \centering
    \caption{High-Density scores obtained for each model. Results in \textbf{bold} highlights the top-performer.}
    \scriptsize
    \resizebox{0.8\textwidth}{!}{\begin{tabular}{l|cccccc|cccccc}
\toprule
 & \multicolumn{6}{c|}{$\alpha$-Precision} & \multicolumn{6}{c}{$\beta$-Recall} \\
 Id & & & & & & & & & & & & \\
 & VAE & E-VAE & EL-VAE & ELD-VAE & LD-VAE & D-VAE & VAE & E-VAE & EL-VAE & ELD-VAE & LD-VAE & D-VAE\\
\midrule
3 & 0.827 & 0.820 & \textbf{0.862} & 0.837 & 0.816 & 0.810 & 0.676 & 0.642 & 0.688 & 0.690 & \textbf{0.698} & 0.653 \\
6 & \textbf{0.909} & 0.871 & 0.734 & 0.723 & 0.707 & 0.738 & 0.010 & 0.007 & 0.025 & 0.025 & \textbf{0.035} & 0.015 \\
11 & 0.676 & 0.647 & 0.494 & 0.607 & 0.925 & \textbf{0.934} & 0.623 & 0.630 & 0.543 & 0.750 & 0.778 & \textbf{0.813} \\
12 & 0.425 & \textbf{0.455} & 0.402 & 0.423 & 0.369 & 0.360 & 0.110 & 0.095 & 0.166 & 0.193 & \textbf{0.218} & 0.165 \\
14 & \textbf{0.270} & 0.264 & 0.231 & 0.216 & 0.256 & 0.249 & 0.265 & 0.265 & 0.322 & 0.362 & \textbf{0.363} & 0.305 \\
15 & 0.744 & 0.701 & \textbf{0.769} & 0.729 & 0.682 & 0.256 & 0.398 & 0.391 & 0.479 & 0.515 & \textbf{0.540} & 0.0190 \\
16 & 0.299 & 0.303 & \textbf{0.305} & 0.252 & 0.240 & 0.227 & 0.172 & 0.150 & 0.258 & \textbf{0.325} & 0.319 & 0.233 \\
18 & \textbf{0.897} & 0.870 & 0.799 & 0.701 & 0.762 & 0.666 & 0.122 & 0.144 & 0.213 & 0.273 & \textbf{0.321} & 0.204 \\
22 & 0.715 & \textbf{0.741} & 0.653 & 0.613 & 0.596 & 0.654 & 0.152 & 0.159 & 0.263 & 0.303 & \textbf{0.306} & 0.195 \\
23 & 0.947 & \textbf{0.976} & 0.948 & 0.975 & 0.953 & 0.886 & 0.471 & 0.462 & 0.474 & 0.478 & 0.483 & \textbf{0.514} \\
29 & \textbf{0.974} & 0.963 & 0.895 & 0.943 & 0.961 & 0.855 & 0.380 & 0.374 & \textbf{0.448} & 0.384 & 0.406 & 0.414 \\
31 & 0.912 & \textbf{0.983} & 0.960 & 0.955 & 0.920 & 0.770 & 0.445 & 0.415 & 0.377 & 0.421 & 0.437 & \textbf{0.472} \\
37 & 0.871 & 0.859 & 0.911 & 0.915 & \textbf{0.927} & 0.857 & 0.288 & 0.300 & 0.351 & 0.389 & 0.358 & \textbf{0.391} \\
38 & 0.764 & \textbf{0.867} & 0.754 & 0.834 & 0.852 & 0.804 & 0.454 & 0.458 & \textbf{0.470} & 0.465 & 0.453 & 0.464 \\
44 & \textbf{0.812} & 0.802 & 0.493 & 0.617 & 0.550 & 0.000 & 0.204 & 0.196 & 0.265 & \textbf{0.273} & 0.261 & 0.000 \\
46 & 0.936 & 0.931 & 0.907 & 0.834 & 0.895 & \textbf{0.939} & 0.428 & \textbf{0.440} & 0.398 & 0.376 & 0.413 & 0.396 \\
50 & 0.923 & 0.910 & 0.896 & 0.920 & 0.911 & \textbf{0.927} & 0.992 & \textbf{0.994} & 0.992 & 0.990 & 0.987 & 0.990 \\
54 & 0.780 & 0.791 & \textbf{0.817} & 0.735 & 0.739 & 0.647 & 0.250 & 0.211 & 0.289 & 0.299 & \textbf{0.332} & 0.306 \\
151 & \textbf{0.927} & 0.866 & 0.868 & 0.854 & 0.898 & 0.919 & 0.151 & 0.159 & 0.172 & \textbf{0.212} & 0.187 & 0.197 \\
182 & 0.884 & \textbf{0.889} & 0.848 & 0.855 & 0.856 & 0.851 & 0.414 & 0.433 & 0.608 & 0.633 & \textbf{0.635} & 0.499 \\
188 & 0.903 & 0.873 & \textbf{0.976} & 0.858 & 0.875 & 0.810 & 0.176 & 0.165 & 0.193 & \textbf{0.205} & 0.195 & 0.152 \\
307 & 0.716 & 0.737 & 0.730 & 0.704 & \textbf{0.739} & 0.726 & 0.013 & 0.012 & 0.012 & \textbf{0.015} & 0.006 & 0.008 \\
458 & 0.454 & \textbf{0.566} & 0.364 & 0.057 & 0.062 & 0.049 & 0.423 & 0.391 & 0.582 & \textbf{0.812} & 0.802 & 0.761 \\
469 & 0.956 & 0.973 & \textbf{0.981} & 0.916 & 0.979 & 0.833 & 0.742 & 0.733 & 0.706 & \textbf{0.747} & 0.726 & 0.677 \\
1049 & 0.901 & \textbf{0.906} & 0.828 & 0.621 & 0.560 & 0.594 & 0.131 & 0.116 & 0.206 & 0.252 & \textbf{0.277} & 0.247 \\
1050 & \textbf{0.923} & 0.903 & 0.922 & 0.755 & 0.687 & 0.691 & 0.144 & 0.134 & 0.228 & 0.302 & \textbf{0.309} & 0.267 \\
1053 & 0.908 & 0.930 & 0.838 & 0.868 & 0.782 & \textbf{0.984} & 0.176 & 0.172 & 0.315 & 0.290 & 0.263 & \textbf{0.452} \\
1067 & 0.831 & 0.839 & \textbf{0.849} & 0.665 & 0.744 & 0.730 & 0.144 & 0.169 & 0.271 & 0.286 & \textbf{0.305} & 0.277 \\
1068 & 0.922 & \textbf{0.934} & 0.814 & 0.795 & 0.741 & 0.736 & 0.191 & 0.193 & 0.289 & 0.343 & 0.357 & \textbf{0.39} \\
1461 & 0.801 & 0.831 & 0.826 & 0.819 & \textbf{0.897} & 0.821 & 0.309 & 0.295 & 0.312 & 0.308 & 0.306 & \textbf{0.321} \\
1462 & \textbf{0.836} & 0.797 & 0.769 & 0.816 & 0.820 & 0.813 & 0.120 & 0.107 & \textbf{0.139} & 0.122 & 0.129 & 0.138 \\
1464 & 0.903 & 0.923 & \textbf{0.938} & 0.892 & 0.872 & 0.820 & 0.276 & 0.328 & 0.364 & 0.355 & 0.378 & \textbf{0.382} \\
1475 & \textbf{0.842} & 0.840 & 0.828 & 0.800 & 0.810 & 0.790 & 0.029 & 0.027 & 0.040 & 0.044 & \textbf{0.046} & 0.041 \\
1480 & 0.938 & 0.914 & \textbf{0.948} & 0.845 & 0.944 & 0.826 & 0.380 & 0.397 & 0.420 & 0.438 & 0.453 & \textbf{0.48} \\
1486 & 0.870 & \textbf{0.877} & 0.781 & 0.799 & 0.737 & 0.838 & 0.067 & 0.063 & 0.153 & \textbf{0.208} & 0.204 & 0.175 \\
1487 & 0.797 & \textbf{0.812} & 0.752 & 0.723 & 0.737 & 0.733 & 0.498 & 0.491 & 0.614 & \textbf{0.627} & 0.624 & 0.578 \\
1489 & 0.871 & 0.853 & 0.860 & 0.818 & \textbf{0.882} & 0.829 & 0.084 & 0.084 & 0.089 & \textbf{0.124} & 0.104 & 0.084 \\
1497 & 0.833 & \textbf{0.861} & 0.711 & 0.609 & 0.605 & 0.680 & 0.040 & 0.045 & 0.070 & 0.085 & \textbf{0.086} & 0.072 \\
1510 & 0.763 & 0.783 & \textbf{0.796} & 0.782 & 0.796 & 0.738 & 0.448 & 0.491 & 0.553 & \textbf{0.593} & 0.567 & 0.549 \\
1590 & 0.861 & 0.852 & 0.762 & 0.829 & 0.813 & \textbf{0.902} & 0.323 & 0.331 & \textbf{0.352} & 0.332 & 0.338 & 0.320 \\
4534 & 0.955 & \textbf{0.961} & 0.912 & 0.926 & 0.864 & 0.926 & 0.387 & 0.386 & 0.422 & 0.415 & \textbf{0.438} & 0.405 \\
4538 & 0.935 & 0.874 & 0.864 & 0.905 & \textbf{0.955} & 0.942 & 0.149 & 0.139 & 0.237 & 0.262 & \textbf{0.267} & 0.237 \\
6332 & \textbf{0.770} & 0.765 & 0.744 & 0.709 & 0.741 & 0.733 & 0.344 & 0.262 & 0.346 & 0.324 & \textbf{0.358} & 0.336 \\
23381 & \textbf{0.974} & 0.971 & 0.864 & 0.924 & 0.835 & 0.917 & 0.450 & 0.463 & \textbf{0.500} & 0.451 & 0.494 & 0.316 \\
40499 & 0.829 & \textbf{0.856} & 0.761 & 0.833 & 0.805 & 0.854 & 0.124 & 0.102 & 0.228 & \textbf{0.254} & 0.233 & 0.150 \\
40668 & 0.814 & 0.829 & 0.835 & 0.804 & 0.803 & \textbf{0.85} & 0.979 & 0.979 & 0.982 & 0.980 & \textbf{0.983} & 0.978 \\
40670 & 0.409 & 0.438 & \textbf{0.556} & 0.406 & 0.430 & 0.324 & 0.668 & 0.676 & 0.675 & \textbf{0.739} & 0.710 & 0.715 \\
40701 & 0.921 & 0.917 & 0.794 & 0.931 & \textbf{0.935} & 0.933 & 0.468 & 0.444 & \textbf{0.503} & 0.500 & 0.490 & 0.473 \\
40966 & 0.508 & \textbf{0.536} & 0.425 & 0.480 & 0.444 & 0.505 & 0.006 & 0.007 & 0.031 & \textbf{0.055} & 0.045 & 0.019 \\
40975 & 0.905 & 0.863 & \textbf{0.98} & 0.887 & 0.900 & 0.814 & 0.991 & 0.988 & \textbf{0.995} & 0.989 & 0.993 & 0.976 \\
40979 & 0.792 & 0.804 & 0.790 & \textbf{0.833} & 0.771 & 0.815 & 0.003 & 0.002 & 0.001 & 0.003 & 0.005 & \textbf{0.006} \\
40982 & \textbf{0.835} & 0.806 & 0.660 & 0.634 & 0.623 & 0.711 & 0.156 & 0.146 & 0.202 & 0.225 & \textbf{0.243} & 0.215 \\
40983 & \textbf{0.989} & 0.866 & 0.934 & 0.928 & 0.923 & 0.930 & 0.412 & 0.434 & 0.434 & 0.442 & \textbf{0.462} & 0.434 \\
40984 & 0.819 & 0.815 & 0.788 & 0.790 & 0.765 & \textbf{0.831} & 0.071 & 0.054 & 0.113 & 0.106 & \textbf{0.115} & 0.077 \\
40994 & 0.758 & 0.823 & \textbf{0.902} & 0.751 & 0.771 & 0.781 & 0.601 & 0.592 & 0.554 & \textbf{0.65} & 0.628 & 0.596 \\
41027 & 0.324 & 0.323 & 0.438 & 0.696 & 0.677 & \textbf{0.936} & 0.234 & 0.232 & 0.291 & 0.481 & 0.480 & \textbf{0.556} \\
\bottomrule
\end{tabular}
}
    \label{tab:highdensity}
\end{table*}

\newpage

\subsection{Machine-Learning Efficiency}\label{subapp:ml}

\begin{table*}[h]
    \centering
    \caption{Machine-Learning Efficiency scores obtained for each model. Results in \textbf{bold} highlights the top-performer.}
    \scriptsize
    \resizebox{0.8\textwidth}{!}{\begin{tabular}{l|cccccc|cccccc}
\toprule
 & \multicolumn{6}{c|}{ML-Fidelity} & \multicolumn{6}{c}{Utility} \\
 Id & & & & & & & & & & & & \\
 & VAE & E-VAE & EL-VAE & ELD-VAE & LD-VAE & D-VAE & VAE & E-VAE & EL-VAE & ELD-VAE & LD-VAE & D-VAE\\
\midrule
3 & 0.906 & 0.919 & 0.928 & 0.944 & 0.931 & \textbf{0.953} & 0.922 & 0.934 & 0.938 & 0.947 & 0.928 & \textbf{0.962} \\
6 & 0.587 & 0.531 & 0.660 & 0.642 & \textbf{0.700} & 0.598 & 0.579 & 0.525 & 0.648 & 0.640 & \textbf{0.694} & 0.592 \\
11 & 0.841 & 0.825 & 0.714 & \textbf{0.873} & 0.825 & 0.841 & 0.825 & 0.825 & 0.714 & \textbf{0.841} & 0.810 & 0.841 \\
12 & 0.925 & 0.910 & 0.945 & 0.935 & 0.895 & \textbf{0.955} & 0.915 & 0.900 & \textbf{0.945} & 0.925 & 0.900 & 0.940 \\
14 & 0.690 & 0.750 & 0.710 & \textbf{0.795} & 0.720 & 0.730 & 0.725 & 0.690 & 0.690 & \textbf{0.755} & 0.710 & 0.700 \\
15 & 0.957 & 0.957 & 0.971 & \textbf{0.986} & 0.929 & 0.500 & 0.957 & 0.957 & 0.971 & \textbf{0.986} & 0.929 & 0.500 \\
16 & \textbf{0.855} & 0.845 & 0.855 & 0.825 & 0.815 & 0.795 & \textbf{0.865} & 0.850 & 0.855 & 0.820 & 0.825 & 0.790 \\
18 & 0.535 & 0.450 & 0.605 & \textbf{0.675} & 0.655 & 0.545 & 0.520 & 0.440 & 0.490 & 0.565 & \textbf{0.645} & 0.455 \\
22 & 0.680 & 0.720 & \textbf{0.745} & 0.695 & 0.720 & 0.705 & 0.670 & \textbf{0.730} & 0.705 & 0.700 & 0.725 & 0.685 \\
23 & 0.595 & 0.595 & 0.595 & \textbf{0.709} & 0.662 & 0.601 & 0.466 & 0.493 & \textbf{0.554} & 0.554 & 0.486 & 0.514 \\
29 & 0.884 & 0.884 & 0.855 & 0.812 & \textbf{0.913} & 0.812 & 0.812 & \textbf{0.870} & 0.812 & 0.768 & 0.870 & 0.797 \\
31 & \textbf{0.830} & 0.780 & 0.790 & 0.760 & 0.820 & 0.770 & 0.690 & \textbf{0.700} & 0.690 & 0.700 & 0.660 & 0.690 \\
37 & \textbf{0.857} & 0.792 & 0.818 & 0.740 & 0.792 & 0.792 & \textbf{0.818} & 0.779 & 0.727 & 0.727 & 0.779 & 0.727 \\
38 & 0.974 & 0.952 & 0.955 & 0.955 & \textbf{0.984} & 0.984 & 0.966 & 0.950 & 0.952 & 0.947 & 0.976 & \textbf{0.981} \\
44 & 0.928 & 0.915 & \textbf{0.939} & 0.928 & 0.939 & 0.644 & 0.889 & 0.868 & 0.892 & \textbf{0.902} & 0.887 & 0.605 \\
46 & 0.940 & 0.906 & \textbf{0.944} & 0.931 & 0.937 & 0.900 & \textbf{0.931} & 0.897 & 0.912 & 0.909 & 0.912 & 0.887 \\
50 & 0.667 & \textbf{0.729} & 0.688 & 0.656 & 0.729 & 0.646 & 0.646 & \textbf{0.708} & 0.667 & 0.656 & 0.708 & 0.625 \\
54 & 0.729 & 0.671 & 0.659 & 0.576 & \textbf{0.741} & 0.612 & \textbf{0.694} & 0.612 & 0.635 & 0.624 & 0.671 & 0.624 \\
151 & 0.831 & 0.837 & 0.842 & \textbf{0.846} & 0.841 & 0.843 & 0.761 & 0.759 & 0.768 & \textbf{0.772} & 0.765 & 0.767 \\
182 & 0.880 & 0.888 & 0.886 & \textbf{0.893} & 0.877 & 0.880 & 0.835 & 0.843 & 0.846 & \textbf{0.857} & 0.848 & 0.838 \\
188 & 0.459 & 0.541 & 0.568 & 0.486 & \textbf{0.622} & 0.500 & 0.432 & 0.486 & 0.595 & 0.365 & \textbf{0.676} & 0.473 \\
307 & 0.485 & 0.495 & \textbf{0.576} & 0.485 & 0.556 & 0.535 & 0.525 & 0.495 & \textbf{0.586} & 0.475 & 0.556 & 0.505 \\
458 & 0.894 & 0.894 & 0.918 & \textbf{0.929} & 0.894 & 0.835 & 0.918 & 0.918 & 0.941 & \textbf{0.953} & 0.918 & 0.859 \\
469 & 0.138 & 0.188 & 0.150 & 0.288 & 0.188 & \textbf{0.362} & 0.175 & \textbf{0.238} & 0.212 & 0.212 & 0.188 & 0.212 \\
1049 & 0.918 & 0.904 & 0.884 & 0.870 & 0.904 & \textbf{0.925} & 0.904 & 0.890 & 0.897 & 0.856 & 0.890 & \textbf{0.911} \\
1050 & 0.936 & 0.936 & 0.936 & \textbf{0.943} & 0.930 & 0.943 & 0.898 & 0.885 & 0.873 & \textbf{0.917} & 0.892 & 0.879 \\
1053 & 0.946 & 0.938 & 0.938 & \textbf{0.958} & 0.950 & 0.910 & 0.807 & 0.807 & 0.803 & \textbf{0.812} & 0.810 & 0.795 \\
1067 & 0.905 & \textbf{0.938} & 0.915 & 0.924 & 0.934 & 0.938 & 0.820 & \textbf{0.863} & 0.820 & 0.839 & 0.829 & 0.834 \\
1068 & \textbf{0.982} & 0.982 & 0.982 & 0.982 & 0.982 & 0.982 & \textbf{0.937} & 0.937 & 0.937 & 0.937 & 0.937 & 0.937 \\
1461 & 0.946 & 0.942 & 0.950 & 0.947 & \textbf{0.954} & 0.943 & 0.892 & 0.888 & \textbf{0.893} & 0.890 & 0.892 & 0.886 \\
1462 & 0.935 & 0.957 & \textbf{0.971} & 0.971 & 0.957 & 0.949 & 0.942 & 0.964 & \textbf{0.978} & 0.978 & 0.949 & 0.957 \\
1464 & \textbf{0.893} & 0.893 & 0.853 & 0.867 & 0.893 & 0.867 & \textbf{0.773} & 0.747 & 0.733 & 0.747 & 0.720 & 0.693 \\
1475 & \textbf{0.536} & 0.472 & 0.520 & 0.498 & 0.505 & 0.534 & 0.436 & 0.407 & \textbf{0.449} & 0.417 & 0.400 & 0.446 \\
1480 & 0.746 & 0.797 & 0.814 & 0.831 & 0.780 & \textbf{0.847} & 0.729 & 0.678 & 0.729 & \textbf{0.746} & 0.695 & 0.729 \\
1486 & 0.929 & 0.956 & 0.945 & 0.956 & \textbf{0.961} & 0.957 & 0.913 & \textbf{0.938} & 0.933 & 0.934 & 0.938 & 0.934 \\
1487 & 0.957 & 0.961 & 0.961 & \textbf{0.972} & 0.965 & 0.972 & 0.937 & \textbf{0.949} & 0.941 & 0.945 & 0.945 & 0.937 \\
1489 & 0.754 & 0.778 & \textbf{0.780} & 0.762 & 0.774 & 0.758 & 0.741 & \textbf{0.762} & 0.741 & 0.738 & 0.758 & 0.741 \\
1497 & 0.914 & 0.885 & 0.874 & \textbf{0.932} & 0.885 & 0.901 & 0.910 & 0.881 & 0.874 & \textbf{0.929} & 0.885 & 0.901 \\
1510 & 0.930 & \textbf{0.965} & 0.930 & 0.965 & 0.947 & 0.965 & 0.965 & 0.965 & 0.965 & \textbf{1.000} & 0.982 & 0.965 \\
1590 & 0.920 & 0.921 & 0.922 & 0.920 & 0.917 & \textbf{0.932} & 0.829 & 0.833 & 0.833 & 0.834 & 0.833 & \textbf{0.838} \\
4534 & 0.953 & \textbf{0.965} & 0.954 & 0.960 & 0.953 & 0.943 & 0.911 & \textbf{0.932} & 0.923 & 0.920 & 0.922 & 0.912 \\
4538 & 0.608 & 0.575 & 0.611 & 0.549 & 0.637 & \textbf{0.65} & 0.460 & 0.437 & 0.456 & 0.455 & \textbf{0.472} & 0.466 \\
6332 & 0.611 & 0.611 & 0.704 & \textbf{0.741} & 0.685 & 0.704 & 0.648 & \textbf{0.722} & 0.667 & 0.704 & 0.685 & 0.667 \\
23381 & 0.700 & 0.660 & 0.640 & \textbf{0.720} & 0.700 & 0.580 & 0.500 & 0.660 & \textbf{0.68} & 0.600 & 0.580 & 0.540 \\
40499 & 0.936 & 0.933 & 0.942 & \textbf{0.951} & 0.945 & 0.927 & 0.940 & 0.938 & 0.949 & \textbf{0.969} & 0.956 & 0.938 \\
40668 & 0.847 & 0.854 & 0.862 & 0.850 & \textbf{0.863} & 0.857 & 0.691 & 0.696 & 0.703 & 0.694 & \textbf{0.707} & 0.701 \\
40670 & 0.897 & 0.912 & 0.903 & 0.912 & \textbf{0.931} & 0.875 & 0.865 & 0.887 & 0.893 & 0.884 & \textbf{0.900} & 0.850 \\
40701 & 0.908 & 0.898 & 0.908 & 0.902 & 0.898 & \textbf{0.918} & 0.870 & 0.864 & 0.874 & 0.868 & 0.860 & \textbf{0.88} \\
40966 & 0.667 & 0.741 & \textbf{0.769} & 0.731 & 0.630 & 0.694 & 0.685 & 0.759 & \textbf{0.769} & 0.759 & 0.648 & 0.676 \\
40975 & 0.717 & \textbf{0.769} & 0.763 & 0.723 & 0.746 & 0.728 & 0.711 & 0.763 & \textbf{0.769} & 0.717 & 0.746 & 0.728 \\
40979 & 0.775 & \textbf{0.805} & 0.755 & 0.745 & 0.700 & 0.620 & 0.770 & \textbf{0.815} & 0.740 & 0.745 & 0.715 & 0.625 \\
40982 & \textbf{0.728} & 0.703 & 0.651 & 0.656 & 0.626 & 0.615 & \textbf{0.667} & 0.651 & 0.590 & 0.590 & 0.595 & 0.595 \\
40983 & 0.969 & \textbf{0.983} & 0.977 & 0.971 & 0.983 & 0.975 & 0.959 & 0.969 & \textbf{0.971} & 0.965 & 0.969 & 0.961 \\
40984 & \textbf{0.848} & 0.823 & 0.805 & 0.831 & 0.784 & 0.801 & 0.831 & 0.805 & 0.805 & \textbf{0.835} & 0.766 & 0.771 \\
40994 & \textbf{1.000} & 0.944 & 0.963 & 1.000 & 0.981 & 0.981 & \textbf{0.963} & 0.907 & 0.926 & 0.963 & 0.944 & 0.944 \\
41027 & 0.716 & 0.713 & 0.717 & 0.748 & 0.771 & \textbf{0.780} & 0.666 & 0.666 & 0.669 & 0.685 & 0.710 & \textbf{0.719} \\
\bottomrule
\end{tabular}
}
    \label{tab:ml_eff}
\end{table*}

\newpage

\section{Representation Similarities}

We present similarities obtained between Transformer representations in the considered VAE variations for the studied datasets below.

\subsection{E-VAE}

\begin{figure}[ht]
    \centering
    \includegraphics[width=\textwidth]{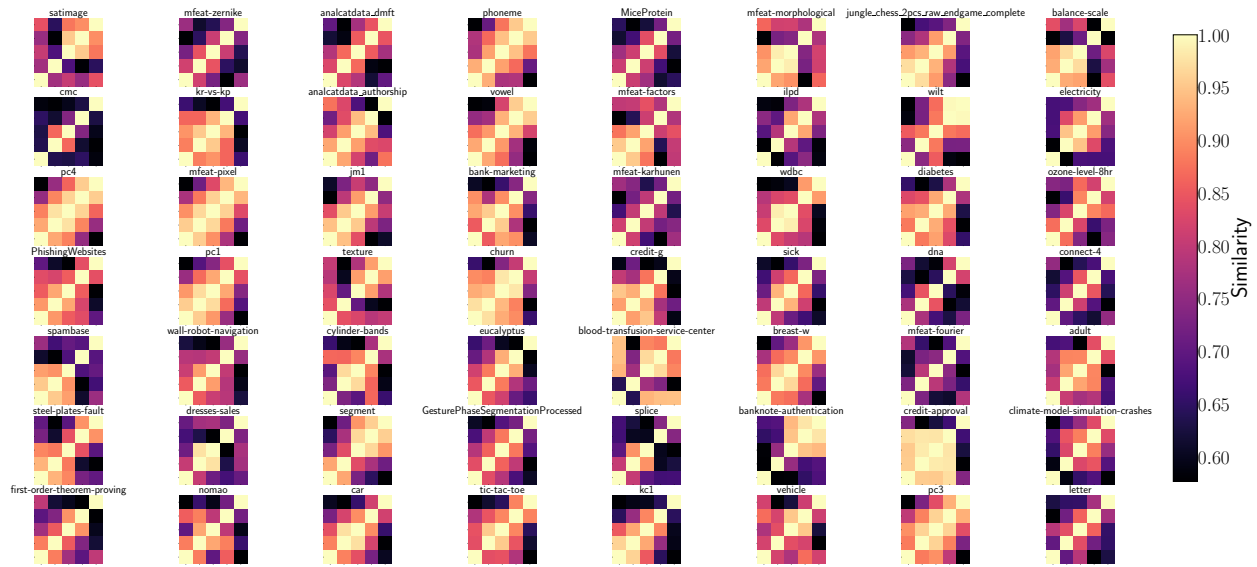}
    \caption{Transformer representation similarities for E-VAE.}
    \label{fig:evae}
\end{figure}

\subsection{EL-VAE}

\begin{figure}[ht]
    \centering
    \includegraphics[width=\textwidth]{plots/all_similarities_ELVAE.pdf}
    \caption{Transformers representation similarities for EL-VAE for the considered datasets.}
    \label{fig:elvae}
\end{figure}

\newpage

\subsection{ELD-VAE}

\begin{figure}[ht]
    \centering
    \includegraphics[width=\textwidth]{plots/all_similarities_ELDVAE.pdf}
    \caption{Transformers representation similarities for ELD-VAE for the considered datasets.}
    \label{fig:eldvae}
\end{figure}

\subsection{LD-VAE}

\begin{figure}[ht]
    \centering
    \includegraphics[width=\textwidth]{plots/all_similarities_LDVAE.pdf}
    \caption{Transformers representation similarities for LD-VAE for the considered datasets.}
    \label{fig:ldvae}
\end{figure}

\newpage

\subsection{D-VAE}

\begin{figure}[ht]
    \centering
    \includegraphics[width=\textwidth]{plots/all_similarities_DVAE.pdf}
    \caption{Transformer representation similarities for D-VAE for the considered datasets.}
    \label{fig:dvae}
\end{figure}

\bibliographystyle{plain}
\bibliography{sup}